\documentclass[lettersize,journal]{IEEEtran}
\usepackage{amsmath,amsfonts}
\usepackage{algorithmic}
\usepackage{algorithm}
\usepackage{array}
\usepackage[caption=false,font=normalsize,labelfont=sf,textfont=sf]{subfig}
\usepackage{textcomp}
\usepackage{stfloats}
\usepackage{url}
\usepackage{verbatim}
\usepackage{graphicx}
\usepackage{cite}
\usepackage{xcolor}
\usepackage{hyperref}
\usepackage{listings}
\usepackage{tabularx}
\usepackage{booktabs}

\begin{document}

\title{
An Edge–Host–Cloud Architecture for Robot-Agnostic, Caregiver-in-the-Loop Personalized Cognitive Exercise: Multi-Site Deployment in Dementia Care
}


\author{Wenzheng Zhao,~\IEEEmembership{Student Member,~IEEE,}
Ruth Palan Lopez,
Shu Fen Wung,
        Fengpei Yuan,~\IEEEmembership{Member,~IEEE}
\thanks{This work has been submitted to IEEE Transactions on Robotics (T-RO) for possible publication.
The authors are with Worcester Polytechnic Institute.
Corresponding author: fyuan3@wpi.edu}
}




\maketitle

\begin{abstract}

We present \textbf{\textit{Speaking Memories}}, a distributed, stakeholder-in-the-loop robotic interaction platform for personalized cognitive exercise support. Rather than a single robot-centric system, Speaking Memories is designed as a generalizable robotics architecture that integrates caregiver-authored knowledge, local edge intelligence, and embodied robotic agents into a unified socio-technical loop. The platform fuses auditory, visual, and textual signals to enable emotion-aware, personalized dialogue, while decoupling multimodal perception and reasoning from robot-specific hardware through a local edge interaction server. This design achieves low-latency, privacy-preserving operation and supports scalable deployment across heterogeneous robotic embodiments. Caregivers and family members contribute structured biographical knowledge via a secure cloud portal, which conditions downstream dialogue policies and enables longitudinal personalization across interaction sessions. Beyond real-time interaction, the system incorporates an automated multimodal evaluation layer that continuously analyzes user responses, affective cues, and engagement patterns, producing structured interaction metrics at scale. These metrics support systematic assessment of interaction quality, enable data-driven model fine-tuning, and lay the foundation for future clinician- and caregiver-informed personalization and intervention planning. We evaluate the platform through real-world deployments, measuring end-to-end latency, dialogue coherence, interaction stability, and stakeholder-reported usability and engagement. Results demonstrate sub-6-second response latency, robust multimodal synchronization, and consistently positive feedback from both participants and caregivers. Furthermore, 
subsets of the dataset can be shared upon request, subject to participant consent and IRB constraints.

\end{abstract}


\section{Introduction}

\IEEEPARstart{A}{lzheimer’s} disease and related dementias (ADRD) affect over 55 million people worldwide, with nearly 10 million new cases each year \cite{who2025dementia}. Beyond cognitive decline and memory impairment, individuals living with dementia often experience social isolation, emotional distress, and reduced engagement in meaningful activities, which significantly impact their quality of life \cite{bemelmans2012socially}.  This growing population also imposes a substantial societal and economic burden. In 2019 alone, dementia was estimated to cost the global economy US\$1.3 trillion, with approximately 50\% (\$651.4 billion) attributed to unpaid care provided by family members and friends. Informal caregivers contribute an average of five hours of daily care, with women accounting for nearly 70\% of this effort. As pharmacological treatments remain limited, non-pharmacological interventions that provide personalized, socially engaging, and emotionally supportive interactions have become increasingly important in dementia care \cite{livingston2020dementia}. In this context, social assistive robots (SARs) have shown strong potential to support the well-being of older adults and individuals living with dementia by fostering engagement, reducing caregiver burden, and promoting emotional connectedness~\cite{deframework,mataric2016socially,fu2025personalizedsociallyassistiverobots}. Among the many forms of interventions, reminiscence therapy~\cite{park2019systematic, otaka2024positive} has been widely adopted in dementia cognitive exercise care. This approach leverages personal photographs, music, and life stories to evoke autobiographical memories, demonstrating positive effects on mood, social connectedness, and quality of life. However, translating this inherently human practice into robotic systems presents significant challenges~\cite{yuan2023cognitive}. Robots must be able to interpret multimodal user inputs, adapt conversations based on individual affective states, and deliver reliable performance in dynamic environments. Although existing SAR systems have made important progress, many remain constrained by rule-based dialogue templates, a limited emotion-awareness, or reliance on onboard hardware with with insufficient capacity for real-time multimodal reasoning~\cite{shi2022toward,hung2019benefits,otaka2024positive,sawik2023robots}. Recent advances in large language models (LLMs) and vision–language models (VLMs) offer new opportunities for creating adaptive and affect-sensitive human–robot interactions~\cite{wang2024lami,stark2024dobby}. yet integrating such models into embodied systems raises significant engineering and methodological obstacles: (i) real-time inference on resource-constrained robot hardware,
(ii) multimodal synchronization under noisy clinical conditions,
(iii) personalization based on caregiver knowledge, and
(iv) privacy-preserving deployment in healthcare settings.

\begin{figure}[t]
  \centering
  \includegraphics[width=0.95\linewidth]{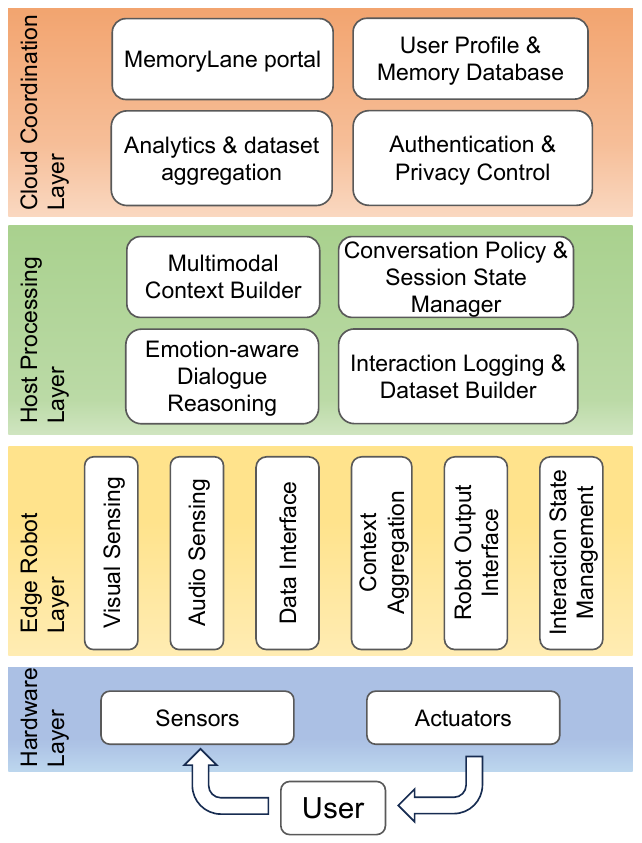}
  \caption{System architecture of the \textit{Speaking Memories} framework, illustrating a layered host–edge–cloud design for emotion-aware reminiscence interaction.
The \textbf{Hardware Layer} interfaces with the user through onboard sensors and actuators.
The \textbf{Edge Robot Layer} performs real-time multimodal sensing, lightweight context aggregation, and robot-side interaction state management, enabling low-latency perception and feedback independent of robot embodiment.
The \textbf{Host Processing Layer} handles multimodal context construction, emotion-aware dialogue reasoning, conversation policy execution, and interaction logging, serving as the core runtime intelligence of the system.
The \textbf{Cloud Coordination Layer} supports caregiver-in-the-loop personalization and long-term operation by managing user profiles and memory databases, authentication and privacy control, and large-scale analytics and dataset aggregation.
This separation of concerns decouples real-time interaction from long-term data management, enabling privacy-preserving deployment, cross-session personalization, and scalable operation across heterogeneous robotic platforms.}
  \label{fig:System_Architecture}
\end{figure}

To address these challenges, we present \textbf{Speaking Memories}, a \emph{robot-agnostic, host–edge multimodal dialogue framework} designed for emotion-aware reminiscence interaction.
As shown in Fig.~\ref{fig:System_Architecture}, the system decouples perception, reasoning, and embodiment through a hybrid architecture: a cloud-based caregiver portal manages biographical knowledge; an edge host performs real-time multimodal inference using VLM/LLM models; and a lightweight robot client delivers speech, display, and nonverbal cues. 
This design enables (i) real-time multimodal reasoning, (ii) plug-and-play deployment across heterogeneous robots, and (iii) privacy-preserving, caregiver-informed personalization. This enhancement design improves flexibility, computational efficiency, scalability, and centralized backend management.

\textbf{Our contributions are as follows:}
\begin{itemize}
    \item 
    {Robot-independent host--edge--cloud architecture for deployable personalized HRI.} We present a deployable system that decouples high-level multimodal perception and reasoning from robot-specific I/O and control, and supports multimodal feedback synthesis and execution (speech, prosody, gaze, gesture, and graphical cues). This separation enables thin, portable robot clients and facilitates scalable deployment of real-time, emotion-aware personalized reminiscence interactions.

    \item Caregiver-in-the-loop personalization pipeline via a structured memory portal. We introduce \textit{MemoryLane}, a cloud-based portal that organizes biographical information, personal media, and dementia-informed attributes into a structured, retrievable memory base. At runtime, the system combines implicit affect inference from visual and prosodic cues with VLM–LLM prompting to generate dementia-appropriate, personalized reminiscence dialogue grounded in user profiles and media. This design enables stakeholder-authored personalization and governance without requiring caregiver presence during each interaction session.

    \item Iterative, multi-phase stakeholder-in-the-loop field evaluation with system-level ablation. We conduct a three-phase evaluation with dementia-care experts, older adults, people with dementia and their caregivers: (i) co-design of a dementia-informed, personalized reminiscence framework; (ii) an initial field deployment revealing latency and turn-taking challenges; (iii) a system-level upgrade that included dual-surface, latency-aware dialog pipeline; and (iv) a second field validation with individuals with dementia to confirm usability and engagement.  

    \item Structured multimodal logging and objective engagement metrics for dementia-oriented HRI. 
    We develop a synchronized multimodal logging format that captures dialogue, affective cues, and model outputs during real-world deployments, enabling reproducible analysis and future model development. Building on these logs, we propose (i) an LLM-based semantic engagement metric and (ii) a VLM-based nonverbal engagement metric to quantify reminiscence interaction quality, and validate both against human coder annotations. Subsets of the dataset can be shared upon request, subject to participant consent and IRB constraints.
\end{itemize}

\section{Related Work}

Socially assistive robotics has evolved along multiple research directions, spanning system design, multimodal interaction intelligence, personalization mechanisms, and domain-specific therapeutic applications. While significant progress has been made within each of these areas, integrating them into deployable, scalable architectures for real-world care settings remains challenging.

To contextualize our contribution, we review prior work along five intersecting dimensions: (i) robot system architectures for social interaction, (ii) edge–cloud partitioning strategies for real-time multimodal pipelines, (iii) stakeholder-in-the-loop personalization mechanisms,  (iv) dementia-oriented SAR as a demanding application domain, and (v) LLM/VLM in Embodied Dialogue.

\subsection{Robot System Architectures for Social Interaction}

Socially assistive robots (SAR) have historically been developed as tightly integrated systems in which perception, dialogue management, and embodiment-specific control are co-designed for particular therapeutic or social tasks~\cite{mataric2016socially, feil2005defining, hung2019benefits, dautenhahn2009kaspar}, however, many SAR systems remain embodiment-bound, with interaction logic deeply coupled to hardware-specific sensing and actuation pipelines.. While such vertical integration enables rapid prototyping within constrained domains, it often limits architectural portability across heterogeneous robot platforms and deployment environments.  Furthermore, most remain tailored to specific robot embodiments or interaction paradigms~\cite{liao2023use, yuan2021systematic}, and few explicitly address robot-agnostic deployment in real-world care environments where hardware, networking conditions, and computational resources vary substantially. Beyond what is mentioned above, real-world social interaction systems must satisfy additional constraints, including real-time responsiveness, multimodal synchronization, cross-session data persistence, and multi-site maintainability. Balancing these constraints while preserving interaction quality remains an open systems-level challenge in socially assistive robotics.

In contrast, our work proposes a robot-agnostic interaction architecture that explicitly decouples multimodal perception and dialogue reasoning from hardware-specific actuation. By separating edge-level sensing and feedback execution from host-level interaction intelligence and cloud-level personalization governance, the proposed framework supports scalable, deployable social interaction across heterogeneous robot platforms and care settings.

\subsection{Edge–Cloud Robotics for Real-Time Multimodal Interaction}

The integration of multimodal perception, language understanding, and action generation in real-time human–robot interaction (HRI) requires computational resources that frequently exceed the capabilities of onboard processors. Distributed and cloud robotics paradigms have therefore explored architectural partitioning strategies that allocate sensing, inference, and control across edge devices and remote servers~\cite{huang2022edge,hu2012cloud,zhang2022distributed}. Edge computing enables low-latency processing close to the data source, while cloud infrastructures support large-scale model hosting, global knowledge sharing, and asynchronous updates. Hybrid edge–cloud schemes have been applied to mobile robotics and collaborative platforms to balance responsiveness with scalability~\cite{huang2022edge,hu2012cloud}. Within multimodal HRI, distributed deployments have supported speech recognition, visual perception, and dialogue generation pipelines~\cite{zhang2022distributed,anari2025privacy,yang2022cloud}. Edge modules typically handle time-critical signal processing and user-facing interaction, whereas cloud services perform computationally intensive inference and long-term data management. These designs improve robustness and privacy by reducing raw data transmission and enabling partial autonomy under network variability. However, existing systems are often optimized for individual perception channels or navigation tasks, and rarely address the coordinated scheduling and synchronization of multimodal streams required for socially assistive dialogue. Real-time conversational interaction introduces strict latency constraints for turn-taking, cross-modal alignment, and session-level state consistency, which are not the primary focus of many navigation- or perception-centric edge–cloud architectures. 

In contrast to task-specific offloading strategies, our work adopts an interaction-centric partitioning principle that explicitly separates sensing, multimodal reasoning, and personalization governance, which across our presented framework adopts a hybrid edge–cloud architecture that unifies local multimodal inference on a Jetson-based embedded unit with a co-located PC server, while also using cloud-hosted dataset curation and coordination. This structured separation supports real-time multimodal dialogue while maintaining scalability, robustness, and privacy considerations in socially assistive deployments.

\subsection{Stakeholder-in-the-Loop Robot Personalization}

Personalization has long been recognized as a critical factor in effective human–robot interaction, particularly in assistive and therapeutic settings where individual preferences~\cite{otake2021cognitive}, cognitive states~\cite{yuan2021systematic}, and life histories shape engagement quality. Prior work in socially assistive robotics has explored adaptive dialogue strategies, user modeling techniques, and reinforcement learning–based personalization~\cite{yuan2021learning} to tailor interaction content and behavior to individual users~\cite{park2019systematic, hsieh2023socially}. However, most existing personalization mechanisms focus on automated adaptation driven by inferred user states or learned policies, with limited explicit involvement from external stakeholders. In real-world care environments, personalization often requires structured oversight from caregivers or clinicians to ensure appropriateness, safety, and alignment with therapeutic goals.  

In contrast, our work embeds stakeholder-in-the-loop personalization within the overall system design, enabling caregiver-guided content curation and longitudinal adaptation through structured data flow across edge, host, and cloud layers.

\subsection{Dementia-Oriented SAR as Target Domain}

Socially assistive robots~\cite{feil2005defining}have been extensively investigated in dementia care and reminiscence therapy, where embodied interaction can promote emotional engagement, cognitive stimulation, and social connection~~\cite{hung2019benefits,dautenhahn2009kaspar,liao2023use}. Reminiscence-based interaction~\cite{yuan2021systematic}, in particular, leverages personally meaningful media to evoke autobiographical memories and sustain engagement. For example, storytelling and photo-based dialogue systems have been implemented to facilitate narrative recall~\cite{otake2021cognitive}, while music-driven interfaces and game-like activities have been used to sustain users' attention~\cite{deframework}. Despite demonstrated therapeutic potential, dementia-oriented SAR systems face distinct deployment challenges. Cognitive variability, fluctuating affective states, and the need for longitudinal adaptation require systems that are both responsive and carefully governed. Furthermore, real-world care settings impose constraints related to privacy, personalization oversight, and consistency that extend beyond laboratory prototypes. 

We therefore adopt dementia-care reminiscence interaction not merely as an application scenario, but as a demanding target domain that exposes the architectural requirements of deployable, stakeholder-governed social robotics systems.

\subsection{LLM/VLM in Embodied Dialogue}
Recent advances in large language models (LLMs) and vision–language models (VLMs) have substantially expanded the perceptual and reasoning capabilities of embodied agents~\cite{2023GPT4VisionSC,driess2023palm}. Pretrained multimodal models enable open-ended dialogue, scene interpretation, and instruction following, reducing reliance on scripted interaction flows and task-specific pipelines. Embodied frameworks that integrate LLM or VLM backbones within robotic platforms have demonstrated the feasibility of grounding natural language interaction in visual perception and physical action~\cite{zitkovich2023rt,huang2022inner}. These developments significantly broaden the expressive capacity of socially interactive robots.

However, while large models provide powerful reasoning primitives, they do not in themselves resolve systems-level challenges related to real-time deployment, architectural partitioning, stakeholder governance, or long-term personalization in care environments. Integrating large-model capabilities into deployable, scalable, and ethically governed social robotics architectures, therefore, remains an open challenge. In this work, LLMs and VLMs function as enabling components within a broader distributed architecture; the focus is on how large-model capabilities can be structured, governed, and deployed within real-world socially assistive systems.

\begin{figure*}[!t]
  \centering
  \includegraphics[width=\linewidth]{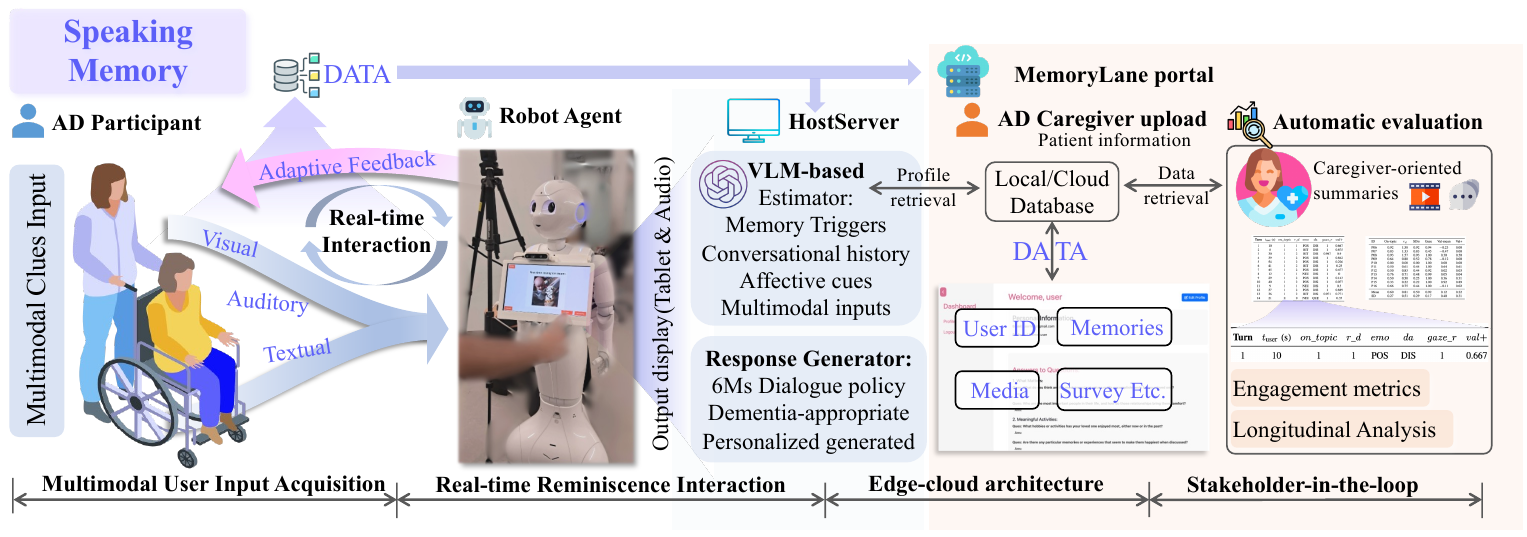}  
  \caption{
    Pipeline of the \textbf{Speaking Memories} platform, a distributed, stakeholder-in-the-loop architecture for adaptive reminiscence interaction. The system comprises three tightly coupled layers.
\textbf{(Left)} A multimodal user input layer acquires visual, auditory, and textual cues from the participant during real-time interaction with an embodied robot agent.
\textbf{(Center)} A local edge interaction server decouples multimodal perception and reasoning from robot-specific hardware, performing emotion-aware dialogue generation using vision–language and large language models conditioned on conversational context, affective cues, and structured caregiver-provided knowledge.
\textbf{(Right)} A cloud–local data and analytics layer supports caregiver-in-the-loop personalization through a secure portal and automated multimodal evaluation, producing engagement metrics, longitudinal interaction summaries, and caregiver-oriented reports.
This closed-loop design enables low-latency, privacy-preserving interaction, scalable deployment across heterogeneous robotic embodiments, and continuous data-driven personalization over repeated sessions.
}
  \label{fig:pipeline}
\end{figure*}

\section{System Architecture}
This section presents the system architecture of \textit{Speaking Memories}, designed to support real-time, emotion-aware reminiscence interaction with embodied social robots. The architecture is structured to meet key requirements for socially assistive robotics in dementia care, including low-latency interaction, privacy-preserving data handling, long-term scalability, personalization, and multimodal grounding across heterogeneous robot platforms.

\subsection{Overview and Design Requirements of the Framework}
This section provides a high-level overview of the proposed \textit{Speaking Memories} framework, designed to facilitate real-time, multimodal adaptive dialogue for reminiscence-focused human–robot interaction. As illustrated pipeline in Fig.~\ref{fig:pipeline}, a distributed, stakeholder-in-the-loop architecture for adaptive reminiscence interaction. This long-term adaptation is enabled by the distributed architecture that maintains a persistent user state across sessions, integrates caregiver-in-the-loop knowledge updates, and performs continuous data-driven refinement of interaction policies based on accumulated multimodal interaction history. The system comprises three tightly coupled layers. The proposed architecture is designed to satisfy key robotics constraints inherent to socially assistive interaction in dementia care. Low-latency turn-taking is required to maintain conversational coherence with cognitively impaired users. Privacy-preserving deployment is necessary to ensure that raw multimodal data remain within local networks. Scalable, robot-agnostic embodiment is required to support deployment across heterogeneous platforms without redesign. Longitudinal personalization must be supported across repeated sessions while maintaining structured caregiver oversight. Multimodal grounding must integrate perceptual signals with coordinated feedback modalities, including speech, visual display, and embodied cues. Clinical safety constraints must be enforced at the system level to prevent inappropriate or cognitively overwhelming responses. These constraints collectively motivate a distributed host–edge–cloud architecture.

\subsection{Domain Constraints for Safe Reminiscence Interaction}
The design of the \textit{Speaking Memories} framework was guided by domain-specific constraints inherent to dementia-oriented human–robot interaction. Prior work~\cite{zhou2021assistant, lu2021effectiveness} in dementia care emphasizes the importance of emotionally supportive communication, reduced cognitive load, and structured conversational scaffolding to facilitate autobiographical recall~\cite{russo2019dialogue, otaka2024positive}. Accordingly, the system prioritizes (i) short, clear utterances; (ii) affirmation and emotional validation; and (iii) gradual narrative prompting grounded in personally meaningful media. These constraints informed both the architectural separation between real-time interaction and long-term knowledge management, as well as the design of the dialogue policy that conditions language generation on user-specific profiles and multimodal context~\cite{nyamathi2024establishing}. Importantly, these design decisions reflect a priori considerations derived from clinical best practices rather than outcomes of system evaluation, which are examined separately in Section~\ref{sec:experiments}.

\subsection{Host–Edge–Cloud Architecture}

The architecture is decomposed into three functional components:
\textbf{(i) Edge Robot Component.} Responsible for direct sensor access and multimodal feedback execution. This layer interfaces with audio, vision, and optional tactile inputs, and produces coordinated speech, visual display, and embodied actuation. \textbf{(ii) Host Reasoning Component.}
Responsible for multimodal fusion, dialogue policy execution, latency-critical inference, and session-level state management. \textbf{(iii) Cloud Knowledge Component.} Responsible for user profile storage, caregiver-mediated knowledge acquisition, and longitudinal coordination across sessions and devices.

The separation of these components enables architectural decoupling between sensing, reasoning, and knowledge governance. Each component and the specific execution process will be described in detail in the Section~\ref{Sec: TechnicalImplementation}.

\subsection{Caregiver-in-the-loop Knowledge Acquisition Pipeline}\label{sec:step0}

Personalization is a foundational requirement in dementia-oriented HRI, as users exhibit diverse life histories, emotional triggers, communication styles, and cognitive profiles.
Without personalization, reminiscence dialogue often becomes generic and fails to evoke autobiographical recall—one of the primary clinical mechanisms underpinning reminiscence therapy~\cite{park2019systematic}. Therefore, our framework incorporates a caregiver-in-the-loop knowledge acquisition pipeline that transforms caregiver-provided biographical material into a structured, robot-readable knowledge base that conditions all downstream multimodal reasoning and dialogue generation. To operationalize this process, we adapt the clinical \textbf{6Ms} framework~\cite{Lopez2023AdvancedComfort}, a widely adopted assessment model in dementia care that captures six key domains of personhood: What Matters, Meaningful activities, Mealtimes, Medications and treatments, Mobility, and Making personal care comfortable. 

The AWS-hosted \textit{MemoryLane} portal manages user authentication, media uploads, and the structured metadata storage.
Before the interaction, caregivers can securely upload photographs, narratives, and survey responses to the portal using their credentials. Information collected across these six domains is used to organize photographs, narratives, and survey responses that reflect the participant’s preferences, habits, and emotional triggers. All media and background knowledge are stored in a structured, person-centric format, allowing for efficient indexing and retrieval from the edge server for personalized reminiscence sessions. This design decouples data management from real-time inference, thereby maintaining privacy compliance while supporting longitudinal analytics and cross-session personalization. During interaction, this knowledge base directly conditions the LLM prompting and conversation policy by supplying:
(i) personalized semantic context for reminiscence triggers, (ii) user-specific affective considerations derived from 6Ms annotations, and (iii) guardrails ensuring dementia-appropriate language. As a result, the robot generates responses that are not only contextually coherent but also personally meaningful and clinically aligned, enabling a level of adaptive reminiscence interaction.

\subsection{Personalization and Policy Abstraction}

To support longitudinal, caregiver-aligned interaction without coupling the dialogue logic to any single robot embodiment, we treat personalization as a first-class architectural concern rather than a post-hoc model prompt. Concretely, the framework maintains a session-level \emph{user state} that is updated at each turn and serves as the sole interface between multimodal sensing and downstream interaction decisions. This state aggregates (i) structured profile attributes curated outside the real-time loop, (ii) within-session conversational context and interaction history, (iii) the currently active reminiscence trigger (e.g., media item and its metadata), and (iv) lightweight estimates of engagement and affect derived from available perceptual signals. Importantly, the state representation is intentionally abstract: it is expressive enough to capture clinically relevant context and user-specific preferences, yet does not assume any particular sensing stack or robot hardware. On top of this representation, the interaction controller is implemented as a policy layer that consumes the fused state and produces \emph{coordinated} multimodal outputs. The policy returns a structured action bundle specifying the response content, display, and non-verbal cues. This separation allows the same interaction logic to be reused across platforms, while leaving the concrete realization of actions to the edge robot component.

Personalization enters the loop through three hooks that are stable across embodiments: (1) \emph{retrieval} of user-specific knowledge and constraints from the persistent profile store, (2) \emph{conditioning} of conversational strategy based on session context and the current trigger, and (3) \emph{adaptation} of pacing and scaffolding in response to inferred engagement signals. To ensure clinical appropriateness, policy outputs are routed through an explicit guard layer that enforces cognitive-load and safety constraints before actuation.

\subsection{Design Trade-offs and Robotics Implications}

The proposed architecture reflects several system-level trade-offs commonly encountered in deployable social robotics.

First, the framework adopts a host–edge partitioning strategy rather than relying solely on onboard computation. Fully onboard systems simplify deployment and reduce network dependencies, but they are typically constrained by limited compute resources and memory capacity, particularly for multimodal reasoning workloads. By separating sensing and actuation at the edge from interaction reasoning at the host layer, the system maintains tight turn-taking latency while allowing more computationally demanding inference to be executed off the robot body. This design preserves embodiment and responsiveness while improving scalability as models and interaction policies evolve.

Second, the system prioritizes a robot-agnostic interaction interface instead of tightly coupling software modules to a specific robot hardware stack. While hardware-specific integrations can yield optimized performance for a single platform, they often limit portability and complicate long-term deployment across sites. The proposed abstraction between perception, reasoning, and actuation enables the same interaction logic to be reused across heterogeneous robot embodiments with minimal modification. From a robotics perspective, this abstraction supports broader deployment of socially assistive interaction systems beyond a single robot model.

Finally, the architecture balances local privacy requirements with the benefits of cloud-based coordination. Dementia-oriented interaction involves sensitive personal data, making it undesirable to transmit raw multimodal signals outside the local environment. The system therefore retains real-time sensing and inference within the local host–edge loop while delegating only structured metadata and long-term knowledge management to the cloud layer. This separation allows personalization and cross-session coordination while maintaining privacy-preserving operation in care environments.

Taken together, these design decisions illustrate how the architecture balances responsiveness, scalability, portability, and safety—key requirements for real-world deployment of socially assistive robotic systems.

\section{Technical Implementation}\label{Sec: TechnicalImplementation}

\subsection{System Setup}

\begin{figure}[!htp]
  \centering
  \includegraphics[width=.99\linewidth]{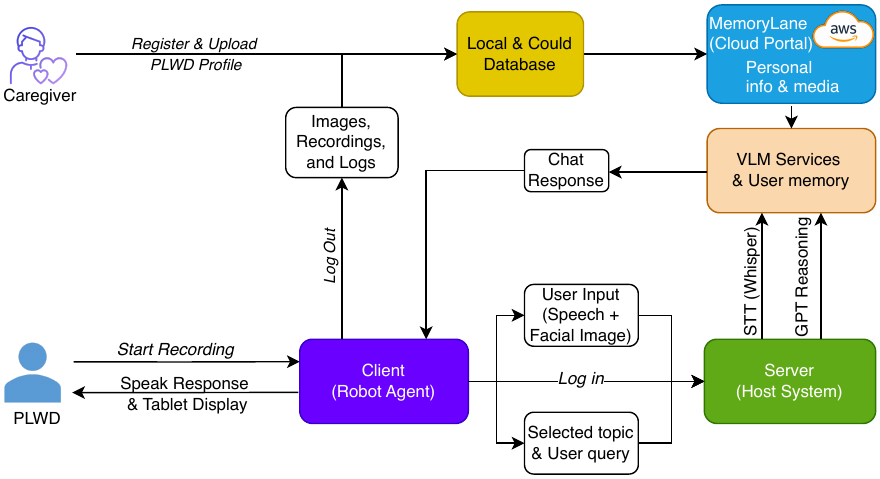}
  \caption{System software flow diagram of the \textit{Speaking Memories} framework for adaptive reminiscence interaction. Colored boxes represent different system components, 
white boxes with borders denote data flow and information exchange. 
Italicized text without boxes indicates user actions, 
while non-italicized text without boxes represents Robot/User-executed actions.}
  \label{fig:software_diagram}
\end{figure}

The detailed software flow of the Agent robot within the Speaking Memories framework is illustrated in Fig.~\ref{fig:software_diagram}. The physical implementation consists of a Jetson Orin NX 16GB unit attached to any robot, a local PC server acting as the primary edge host, and an AWS-based cloud portal.
For the edge-layer implementation, we evaluated two robot-side configurations on two Pepper-based platforms:
(1) Pepper-native configuration  (\textbf{Robot I}): the system used Pepper’s internal computer and embedded sensing suite—including its built-in microphone array, integrated speakers, and head-built-in RGB camera—without any external devices. (2) Jetson-edge configuration (\textbf{Robot II}): a Jetson Orin NX module was externally attached to the robot, together with a ReSpeaker USB microphone array, a JBL Clip 3 speaker, and an Intel RealSense D435i RGB-D camera to provide synchronized audio–visual acquisition and GPU-accelerated local processing. 
The PC server is designed to operate across heterogeneous computing environments, including Windows, macOS, and Ubuntu platforms, all of which have been verified to support the full execution of the framework.  
For laboratory development and benchmarking, the primary setup employs an AMD Ryzen 7950 CPU, an NVIDIA RTX A6000 GPU, and 64 GB of RAM running Ubuntu 22.04 LTS.  
For mobile or field trials, the system was also deployed on a Dell Pro 14 Plus laptop with 16 GB of RAM, running the same Ubuntu 22.04 environment, demonstrating portability and consistent runtime performance across different hardware configurations.
All components communicate over a secure local Wi-Fi or Ethernet network through HTTPS and SSH protocols. A Flask-based RESTful API manages asynchronous data exchange between modules, handling POST/GET requests for audio, visual, and text payloads.
All software modules run on Ubuntu 22.04 with Python 3.8, and the robot interfaces are bridged through the Robot API for low-latency command execution and motion control.

\subsection{Host System Design}
The host system is implemented as a Flask-based web distributed server that handles user interactions, data processing, and robot coordination through an event-driven architecture. This server manages both front-end and back-end logic, offering a browser-based interface for image selection and recording control.  It interfaces with the embedded robot unit using HTTP APIs and SSH-based data exchange. Command signals, such as \texttt{start-recording}, \texttt{stop-recording}, and \texttt{speak-response}, are transmitted from the host to the embedded robot unit via lightweight, JSON-based API calls. Once commands are executed, the embedded unit asynchronously transfers recorded audio and images back to the host through SSH for file transmission, using a dedicated endpoint for state acknowledgment. Captured images are stored locally and organized by dialogue turns. For each turn, up to three images are selected —sampled at 2.0-second intervals— to represent the visual context. These images are encoded and, together with the transcribed user utterance (via Whisper), are passed to a hosted instance of GPT-4o. This model serves as the backbone, depending on computational resources. To generate dementia-appropriate, emotion-aware, and personalized responses, GPT-4o is guided by a structured prompt that incorporates several key elements: (1) the user’s personal information from the \textit{MemoryLane Portal}; (2) the current memory trigger; (3) user facial images to gauge affective cues; (4) the user’s latest response; (5) conversational strategies for reminiscence facilitation; and (6) adjustable agent–user conversation history. The output response is sent simultaneously to the robot for vocalization, displayed as memory trigger, and rendered on the web interface. If a repeat request is made, the server retrieves the last response from memory and reissues the \texttt{play-response} command to the robot.

In summary, the host provides asynchronous coordination of multimodal input and output streams, using an SSH-based pipeline for reliable retrieval of audio and images combined with acknowledgment signaling. Real-time transcription from Whisper  is integrated with GPT-4o dialogue generation, while visual input sampling provides lightweight affective grounding. A memory-based repeat-response mechanism ensures the robustness and continuity of the dialogue, allowing the system to handle interruptions smoothly.

\subsection{Edge Client System Design}

To enable robot-agnostic deployment, we implement the edge client under two configurations that share the same host--cloud architecture and dialogue pipeline, while differing only in edge-level execution:

\textbf{Robot I (Embedded Configuration):} a Pepper-native setup that relies solely on the robot’s onboard computing, sensing, and control interfaces.

\textbf{Robot II (Edge-Augmented Configuration):} a Pepper-based system augmented with an external edge client, implemented on a Jetson Orin NX, which provides enhanced multimodal processing and system orchestration.

Importantly, both configurations preserve identical high-level reasoning, dialogue generation, and cloud coordination mechanisms. This design isolates the role of the edge layer and enables direct evaluation of portability across heterogeneous robot embodiments.

\paragraph*{System Architecture}
In both Robot I and Robot II, the edge layer acts as the multimodal controller and communication gateway between the physical robot and the Flask-based host server.  
The system follows an event-driven loop that polls the host at 0.05\,s intervals for conversation updates, recording triggers, and control signals, ensuring sub-100\,ms response latency between user actions and system feedback. In Robot II, the edge client is deployed on a Jetson Orin NX, which provides GPU acceleration for real-time audio--visual processing. In contrast, Robot I executes the same functional pipeline using the robot’s onboard resources, with reduced computational capacity but identical interaction logic.

Across both configurations, three primary feedback channels are supported:
(1) \textit{speech output} via neural text-to-speech with adaptive playback speed,
(2) \textit{visual display} via the robot tablet or an attached screen, and
(3) \textit{non-verbal behaviors} including gestures and timing coordination aligned with speech output.

\paragraph*{Audio and Image Capture}
Both Robot I and Robot II support synchronized multimodal sensing.  
Audio is captured using a microphone array (e.g., ReSpeaker in Robot II or onboard microphones in Robot I), recorded at 16\,kHz, and processed asynchronously to avoid blocking control execution.  
Recording sessions are triggered by host state signals and stored locally before being uploaded via the \texttt{/audio\_ready} API. In parallel, visual observations are collected using an RGB-D camera at a 640\,$\times$\,480 resolutio (e.g., RealSense D435i in Robot II or onboard cameras in Robot I).  
Image frames are sampled at approximately 0.5\,Hz and indexed by dialogue turn, allowing the host to retrieve aligned visual context through the \texttt{/image\_ready} endpoint.

\paragraph*{Conversation and Display Management}
At session initialization, both configurations reset the conversation state, launch the therapy interface, and begin asynchronous polling for generated responses.  
Language model outputs are streamed from the host, converted into speech, and rendered simultaneously on the display. Repeat commands dynamically adjust playback speed (10\%--25\% slower) to improve accessibility, while maintaining continuous multimodal capture.  
A browser-based interface ensures a minimal and distraction-free user experience, exposing only the active memory cue and essential controls (\textit{Recording}, \textit{Repeat}, \textit{Home}).

\paragraph*{Robustness and Portability}
The edge client follows a modular lifecycle design (\texttt{onLoad}, \texttt{onStart}, \texttt{onStop}, \texttt{onUnload}) to ensure robust operation under device failures or network interruptions.  
In Robot II, this logic is executed on the external edge module, while in Robot I it is mapped onto the robot’s onboard execution framework. Crucially, the edge layer interacts with the robot through lightweight control APIs, issuing only high-level motion or gesture commands.  This abstraction decouples embodiment-specific control from perception and reasoning, allowing the same architecture to be deployed across different robotic platforms with minimal modification.

\paragraph*{Summary and Experimental Role}
This dual-configuration design serves not only as the execution backbone for real-time multimodal interaction, but also as a controlled experimental mechanism for evaluating robot-agnostic deployment.

In Section~\ref{sec:Quantitative_Outcomes}, we compare Robot I and Robot II while keeping all higher-level modules unchanged.  
This setup isolates the contribution of the edge architecture and demonstrates that the proposed framework maintains consistent interaction quality, latency characteristics, and system robustness across heterogeneous robot embodiments.

\subsection{Emotion-Aware Personalized Conversational Pipeline}
The \textit{Edge Robot Layer} handles perception and interaction through microphones, speakers, and a display interface.  It is connected to a Jetson platform for computing. The \textit{Host Layer}—includes a Jetson-based embedded unit and a co-located PC server—performs multimodal signal processing, affect inference, and dialogue generation driven by LLMs in real time. The \textit{Cloud Layer}, hosted on AWS, manages user authentication, dataset storage, and cross-device synchronization. This modular design decouples perception and reasoning from robot-specific hardware, allowing for  deployment across heterogeneous platforms with minimal reconfiguration. The functional workflow of the framework is organized into three stages: data management, multimodal acquisition, and adaptive human–robot interaction. Each stage is detailed in the following subsections (Steps 0–II). Step~0 (see Section~\ref{sec:step0} \& ~\ref{cloud}) manages user profiles and reminiscence media through the AWS \textit{MemoryLane} portal; Step~I captures user multimodal input; and Step~II performs adaptive dialogue generation on the local edge server. 

\subsubsection{Multimodal Acquisition (Step I)}

In this initial stage, multimodal acquisition is performed to sense real-time context-aware and emotion-aware user-centered engagement during the reminiscence session. To enable such interaction, the system employs a multimodal sensing pipeline that captures visual, auditory, and textual cues from the user in real time. These input streams are collected through the Jetson-based embedded unit that connects multiple multimodal acquisition devices and transmits data to the host system for downstream processing.

\paragraph{Visual Cues Acquisition}

Visual data are captured using an Intel RealSense D435i camera to estimate facial expressions during the reminiscence session. 
Once the participant begins to speak, the camera periodically records still images at a default interval of 2.0\,s. 
These frames serve two primary purposes: (i) they provide coarse-grained facial expression cues that help infer the user’s emotional state; and (ii) they supply visual context that grounds the dialogue in the surrounding environment.  The capture interval can be adjusted to accommodate varying interaction scenarios and system response requirements, allowing the framework to balance temporal resolution with computational efficiency.

\paragraph{Auditory Cues Acquisition}

Audio is captured through the onboard microphone array and locally sampled at a fixed rate to synchronize it with visual frames.  
Each recording segment is stored on the Jetson device and securely transferred to the host server via an SSH protocol.  
The audio stream is then transcribed using the Whisper~\cite{radford2023robust} automatic speech-recognition (ASR) model and subsequently analyzed for prosodic emotion cues such as pitch and intensity.  The resulting transcripts and prosodic features constitute essential metadata for the LLM input prompt, enabling alignment with multimodal context and supporting the generation of engagement-sensitive dialogue responses.

\paragraph{Textual Cues Acquisition}

Optional textual input is processed through the tablet interface when participants type responses or select predefined options.  
This modality serves as a fallback mechanism for users who are unable or unwilling to speak, thereby enhancing the robustness and inclusivity of the interaction framework.  
In addition, the textual interface is used by developers and researchers as a diagnostic tool for testing dialogue logic and evaluating system performance.  
All input streams are subsequently fused on the edge server to form a unified contextual embedding for downstream dialogue generation.

\subsubsection{Reminiscence Interaction with Agent (Step II)}
This stage integrates the fused multimodal embeddings into an adaptive dialogue loop designed for emotion-sensitive response generation.  
When the system receives user input, it engages in a real-time reasoning cycle that combines multimodal perception with LLM inference.  
Facial frames are embedded as visual tokens and processed by GPT-4o, which performs implicit emotion inference as part of its reasoning process.  
The model attends to text, audio, and conversational history to infer the user’s affective state and generate responses that are contextually coherent and emotionally aligned, without relying on explicit emotion labels.  
The generated text is synthesized through the Jetson-based text-to-speech (TTS) module, vocalized, and simultaneously displayed on the robot’s tablet interface, forming a closed feedback loop between perception and dialogue execution.  
If the user requests repetition or clarification, the most recent response is retrieved from local memory, ensuring continuity, robustness, and a natural conversational flow.

\subsubsection{Personalized Conversation Policy}

To enable dementia-appropriate, emotionally aligned reminiscence dialogue, we design a personalized conversation policy that conditions all LLM prompting on three categories of user-specific knowledge: (i) the structured biographical profile collected through the MemoryLane portal, (ii) the selected reminiscence media (e.g., personal photos and narratives), and (iii) real-time affective cues extracted from visual and auditory modalities. This policy serves as the core mechanism through which the robot adapts dialogue content, tone, and conversational depth to each individual participant. 

At the beginning of each interaction turn, the system retrieves the user’s 6Ms-derived profile—covering personal information, meaningful activities, emotional triggers, and caregiving preferences—and binds these attributes to a structured LLM prompt template.
Next, the system incorporates context-dependent features: the currently displayed memory trigger, the user’s most recent utterance, and implicit affective estimates derived from facial expression cues and prosodic patterns.
These inputs jointly determine the conversational strategy the robot should take (e.g., supportive, elaborative, affirming, or guiding), as well as the level of narrative scaffolding appropriate for the user’s cognitive condition. These design choices reflec communication principles established by dementia-care experts, ensuring that the conversational style remained clinically appropriate, easy to follow, and supportive of autobiographical recall. The LLM (GPT-4o) then generates a response that is grounded simultaneously in the personalized knowledge base and the immediate multimodal context. This integration transforms the LLM from a generic dialogue generator into a media-grounded, person-centered social agent capable of sustaining autobiographical recall while adhering to dementia communication best practices. This mechanism was essential in enabling coherent, emotionally attuned, and clinically safe conversations across heterogeneous users and memory-care environments. 

Together, these three layers realize a fully integrated, adaptive, and privacy-preserving framework for emotion-aware reminiscence dialogue.

\subsection{Data Management and Cloud Coordination} \label{cloud}

All interaction sessions were automatically logged through an integrated data pipeline connecting the robot client, the local host server, and the AWS-based \textit{MemoryLane} portal. The system recorded synchronized multimodal streams, including (1) user audio waveforms, (2) user speech transcriptions, (3) user facial snapshots, (4) robot dialogue outputs, and (5) system timing information.  
Each data modality was timestamped at acquisition and assigned a session-specific identifier for cross-referencing and temporal alignment.

\paragraph*{Local Data Handling}
During runtime, raw sensor data and intermediate model outputs are cached on both the Jetson device and the host PC for immediate access by the dialogue engine.  
All raw sensor data remain local and are never transmitted to any public or third-party platforms to ensure full data privacy.  
A local scheduler synchronizes data transfer between modules to prevent I/O blocking during active dialogue and automatically retries upon network interruption.

\paragraph*{Cloud Coordination Layer}
The AWS-hosted \textit{MemoryLane Portal} serves as the coordination layer linking long-term user profiles with real-time interaction logs.  
It manages authentication via AWS Cognito, media storage through S3 buckets, and metadata indexing using an RDS MySQL database.  
Caregivers or researchers can securely upload new reminiscence materials, edit user attributes derived from the 6M framework, and review aggregated analytics.  
The portal exposes RESTful endpoints for dataset retrieval and cross-session synchronization, allowing the host server to pre-load user data prior to each interaction.  
All communications are encrypted via HTTPS, and cloud permissions adhere to a least-privilege policy to ensure data confidentiality.

\paragraph*{Synchronization and Logging}
\label{sec:data}
At the end of each session, the host automatically packages multimodal meta-embeddings, conversation transcripts, and associated multimodal metadata into a structured JSON log, while all raw media remain stored locally to maintain privacy compliance (Fig~\ref{fig:dataset}).  
These datasets are timestamped to support downstream analysis and future model development (e.g., fine-tuning domain-specific vision–language models and enabling personalization).
Additionally, a GoPro camera continuously records the entire experiment session onsite as an external validation device to verify the integrity and synchronization of multimodal data or for extended offline analysis.

\begin{figure}[t]
  \centering
  \begin{lstlisting}[basicstyle=\ttfamily\footnotesize,
                     columns=fullflexible, keepspaces=true,
                     xleftmargin=0.12\linewidth, xrightmargin=0.12\linewidth]
dataset/
|-- {user_id}_{conversation_time}/
|   |-- processed_data.json   
|   |-- data.json             
|   |-- images/               
|   |   |-- image_1_1.jpg
|   |   \-- ...
|   \-- recordings/           
|       |-- recording_1_1.wav
|       \-- ...
|-- {user_id}_{conversation_time}/
  \end{lstlisting}
  \caption{Structured dataset format. 
Each session folder contains: (i) \texttt{processed\_data.json} (Includes the conversation content between the participants and the robot agent,  along with the trigger selected by the participant), 
(ii) \texttt{data.json} (raw logs of the session), (iii) \texttt{images/} (Images of the participant), and 
(iv) \texttt{recordings/} (audio files of the participant). Each folder corresponds to one user session and contains metadata,
  raw conversation logs, images, and audio recordings for downstream
  multimodal VLM fine-tuning.}
  \label{fig:dataset}
\end{figure}

\paragraph*{Scalability and Compliance}
This hybrid design supports scalable multi-user deployment across different sites while preserving user privacy and minimizing cloud dependency during active interactions.  
By separating data storage, inference, and analytics across the Jetson edge client, the host PC, and the cloud coordination portal, the framework achieves real-time performance with transparent data provenance and secure lifecycle management.  
As a result, the architecture enables repeatable experimentation, longitudinal study analysis, and integration with future large-scale robotic trials.

\section{Experiments}
\label{sec:experiments}

\subsection{Study Design and Objectives}

This experimental study was designed to evaluate the proposed \textit{Speaking Memories} framework in real-world reminiscing interaction scenarios, focusing on both technical robustness and user experience.  
The evaluation used a mixed-method design, combining quantitative system-level metrics with qualitative user and caregiver feedback.

The primary objectives were threefold:  
(1) to assess system usability and interaction experience using two complementary questionnaires: (i) a participant self-report survey completed by older adults reflecting usability, engagement, and emotional connectedness experienced, and (ii) a caregiver-reported observational survey reflecting caregivers’ perceptions of patient engagement and emotional responses during reminiscence sessions;
(2) to assess the technical reliability and latency performance of the multimodal perception–dialogue pipeline under real-time operating conditions;  
(3) to evaluate the contextual coherence and affective appropriateness of dialogue responses generated by LLM.  To achieve these goals, the study was conducted in three phases. 

\subsection{Study Phases and System Configurations}
To reflect the iterative engineering process required for dementia-appropriate deployment, the evaluation was conducted in three phases, progressing from formative feedback to field feasibility and finally to clinical deployment with people living with dementia. Two robot configurations were used: Robot I and Robot II, as described in Section~\ref{Sec: TechnicalImplementation}.

\textbf{Phase I – Formative evaluation (Senior Center I, Robot I)} : A pilot evaluation was conducted with two dementia-care experts and older adults at a local senior center to collect formative feedback on system usability and clinical applicability. Experts noted that the robot supported session flow, encouraged participation and engagement, and reduced caregiver facilitation demands. Feedback also emphasized the importance of personalization, privacy sensitivity, and technical robustness for long-term adoption. Based on these observations, we iteratively refined the interface and interaction flow prior to the formal on-site studies. 

\textbf{Phase II – Field feasibility (Senior Center II, Robot I)}: To validate the system’s end-to-end stability and usability in real-world environment prior to dementia deployment, we conducted structured experimental sessions with healthy older adults in a different local senior center. Participants completed the full interaction protocol to verify that the system functioned reliably across complete end-to-end use. This phase served as a deployment readiness check, verifying that the full protocol could be completed reliably (e.g., no major latency or interruption issues) after the Phase I refinements.

\textbf{Phase III – Dementia-care deployment and configuration comparison (Assisted Living and Memory Care, Robot I \& Robot II)}: We deployed the system in assisted-living/memory-care facility using both robot configurations (Robot I and Robot II) and conducted full experimental sessions with people living with dementia. This phase enabled a systematic comparison of real-world performance under identical clinical constraints and provided quantitative and qualitative evidence regarding robustness, responsiveness, and interaction quality in dementia-care settings. 

In Phase III, participants interacted with the robot using personalized memory triggers uploaded through the \textit{MemoryLane} portal. Participants, including both persons with dementia and their and caregivers, completed standardized post-session surveys, and multimodal interaction data—including speech, video, and system logs—were recorded for further quantitative analysis.

\textbf{Study Administration}: Phase II and Phase III were conducted between July and September 2025 at partnering senior care centers, memory care center, and assisted living center. The study protocol was approved by an Institutional Review Board (IRB), and informed consent was obtained from all participants. Sessions lasted 5–30 minutes and was facilitated by trained researchers and care staff.

\subsection{Participant Selection Criteria}
Participants were recruited from local senior centers and memory care centers between July and September 2025. Seventeen individuals completed study sessions (6 older adults in Phase II and 11 people living with dementia (PLWD) in Phases III). Per the IRB-approved protocol requiring written consent, data from five older-adult individuals were excluded from analysis due to missing signed consent documentation. All results reported in this manuscript therefore reflect 12 participants (1 healthy older adult and 11 PLWD). If additional sessions were used only for system debugging or engineering validation (not human-subject analysis), we explicitly exclude them from the participant counts in Results.

Participants were identified and referred by caregiving staff familiar with their medical history and cognitive status. Eligibility was assessed using a predefined clinical screening procedure administered by caregiving staff.

\textbf{Inclusion Criteria}:
(1) aged between 50 and 90 years;  
(2) able to understand and communicate in English; 
(3) Either no cognitive impairment (for \textit{Phase II}) or a clinical diagnosis of mild cognitive impairment or early-stage dementia with a Clinical Dementia Rating (CDR) score of 1, or ; and  
(4) no major uncorrected visual or auditory impairments that would interfere with robot-mediated communication.
Exclusion criteria: (1) diagnosis of a non-Alzheimer’s dementia subtype; (2) Clinical Dementia Rating (CDR) score outside of 1; (3) presence of severe behavioral or communicative challenges; (4) unstable medical conditions that would interfere with robot-assisted interaction; (5) lack of personal or meaningful media content (e.g., photographs, music, videos) for reminiscence-based engagement; or (6) absence of a primary caregiver who was willing and able to participate in the study process. 

Because both primary users (persons with dementia) and caregivers provide complementary perspective, the target population for our study included both people living with dementia and their primary caregivers. Both users with dementia and caregivers were invited to complete follow-up surveys to assess usability, emotional impact, and perceived engagement from their own point of view.

\subsection{Experimental Procedure}

\begin{figure}[t]
  \centering
  \includegraphics[width=0.9\linewidth]{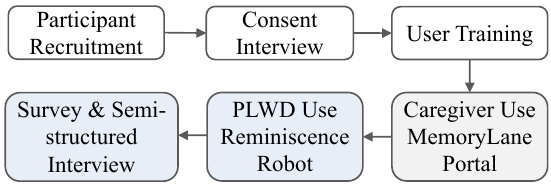}
  \caption{Flowchart of the experimental procedure.}
  \label{fig:procedure}
\end{figure}

\begin{figure}[t]
  \centering
  \includegraphics[width=0.99\linewidth]{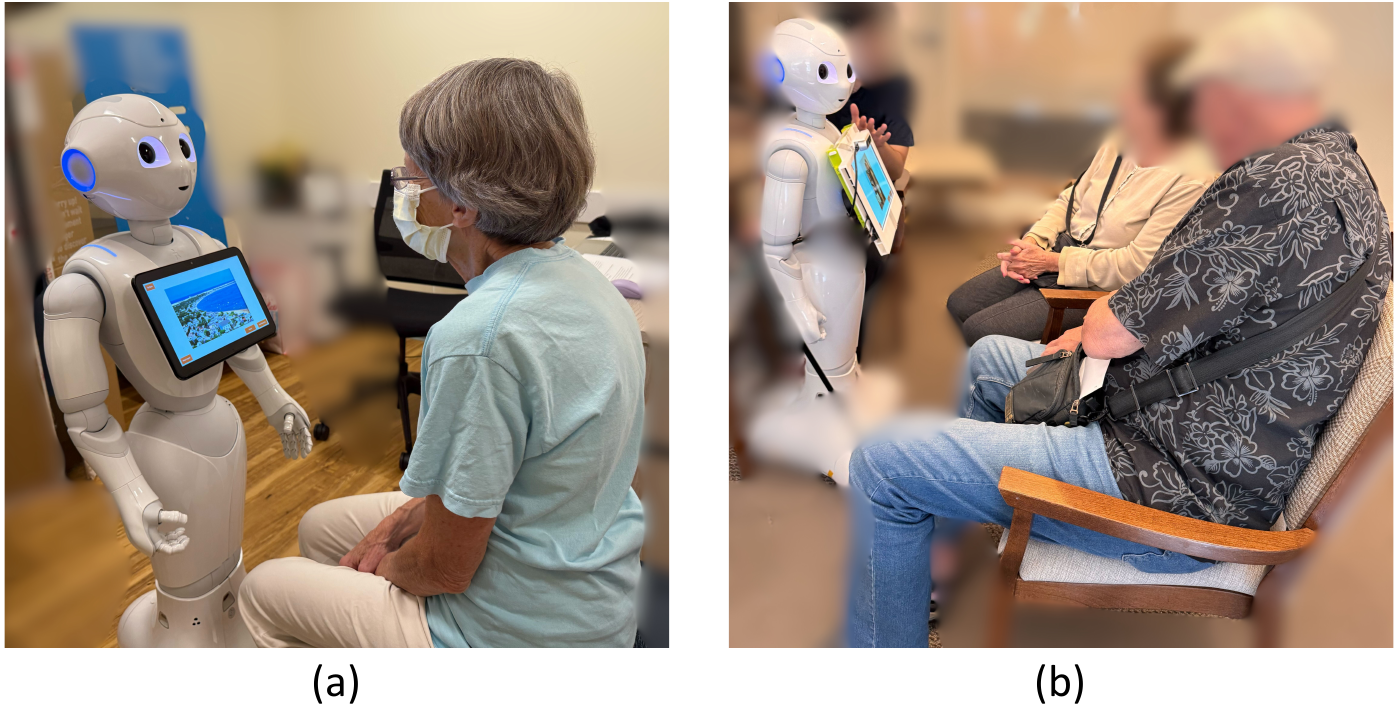}
  \caption{
  Real-world deployment of the \textit{Speaking Memories} framework using two different hardware configurations at an elder care facility.
  (a) The left deployment used \textbf{Robot I}, relied soly on \textbf{ Pepper's native hardware}, using its onboard microphone, camera, and tablet for multimodal interaction.
  (b) The right deployment used \textbf{Robot II}, a \textbf{Jetson-based edge client}, where sensing and feedback were executed on an external Jetson Orin NX unit attached to the Pepper robot.
  Although both configurations delivered personalized reminiscence dialogues, the Jetson setup achieved substantially lower latency and greater modularity by offloading computation to the edge server.}
  \label{fig:onsite}
\end{figure}

Participants were recruited through flyers, outreach videos, and email invitations distributed to partnering senior centers, memory care centers, and assisted living centers. Initially, healthcare staff screened interested individuals for eligibility, after which both people living with dementia (PLWD) and their primary caregivers received IRB-approved consent forms. All sessions were conducted on-site in a quiet, well-lit room familiar to participants to minimize anxiety and ensure ecological validity. The experimental procedure is illustrated in Fig.~\ref{fig:procedure}. Upon enrollment, individuals who met the inclusion criteria were invited to attend an on-site session. 

Prior to each session, a trained research staff member distributed informed consent forms to both the participant with dementia and their primary caregiver, ensuring that both parties fully understood the study’s goals, data collection procedures, and confidentiality measures. After obtaining the written consent forms, participants or their caregivers were guided to log into the \textit{MemoryLane} portal and upload personalized reminiscence materials, typically consisting of photographs and  short notes in response to structured questions derived from the 6M framework (\textit{Meaning, Motivation, Mealtime, Mobility, Medications,} and \textit{Make Comfortable}).  
These materials were securely stored on the cloud portal and preloaded onto the local host before interaction.  
The host system verified data integrity and synchronized the multimedia content with the Jetson-based edge client connected to the robot.  
Once the setup was complete, the participant was seated approximately 0.8~m in front of the robot, facing the tablet display and camera, as shown in Fig.~\ref{fig:onsite}, which shows both configurations deployed at two elder-care facilities. 

Then the participant with dementia logged into the \textit{Speaking Memories} system using the tablet interface. Each session followed a  semi-structured dialogue format designed to balance dementia domain safety control with a natural conversation flow.  
The robot initiated the interaction with a greeting and a brief explanation of the session's purpose.  
Participants selected one of their uploaded memory triggers via the tablet interface.  
Upon selection, the system captured facial snapshots and vocal input for affect inference, then generated a personalized prompt using the GPT-4o model.  
Participants responded verbally after pressing the “Start Recording” button, activating the audio–visual sensing pipeline.  
The Whisper system transcribed speech in real time, visual frames were processed for affective cues, and the LLM generated contextually coherent and emotionally aligned responses. These responses were vocalized through the robot’s text-to-speech (TTS) module and displayed on its screen.  
The dialogue continued for multiple rounds for each memory trigger, until the participant requested to move to the next memory trigger.  
Session duration ranged from 5 to 30 minutes, depending on participant engagement. Throughout the interaction, all multimodal streams—including speech, transcripts, visual frames, timestamps, and generated responses—were automatically logged by the host server.  
A researcher and the participant’s caregiver remained present for safety observation but did not intervene unless there were technical or usability issues.
Environmental parameters, such as lighting and background noise, were kept consistent across sessions to ensure comparability.  
A separate GoPro camera recorded the entire session for validation of synchronization and behavioral annotation. 

Post-session surveys were administered to both participants with dementia and caregivers to capture perceived usability, interaction experience, and perceived system performance during reminiscence sessions.   
Informal interviews were also conducted to collect open-ended feedback on comfort, perceived empathy, and willingness to use such systems in future sessions.  
All responses were anonymized and linked to session identifiers for quantitative and qualitative analysis. Measures included (i) participant-reported usability and engagement using the User Experience Questionnaire (UEQ), and (ii) caregiver ratings of dialogue coherence and affective appropriateness.
and (3) expert qualitative observations focusing on the responsiveness and naturalness of interaction. All study data (logs, recordings, and survey responses) were transferred to IRB-approved secure storage after each experimental session.

\subsection{Evaluation Metrics and Analysis}

\subsubsection{System Performance Metrics}
One of our objective evaluations focused on the system performance, including runtime performance,dialogue quality, and system robustness.  
\textit{End-to-end latency} was computed as the time interval from the end of the user speech to the completion of the robot’s spoken response. This latency was averaged across all dialogue turns within each session. The \textit{dialogue coherence rate} was measured by two independent raters who labeled each system utterance as contextually relevant or irrelevant based on  the preceding user input and conversation history. The final score was calculated by the proportion of relevant responses.  
The \textit{interaction completion rate} was defined as the percentage of sessions that proceeded without technical interruptions or manual intervention from researchers.  
All quantitative logs were automatically extracted from the synchronized dataset described in Section~\ref{sec:data}, and each metric was additionally verified through manual inspection to ensure accuracy. The objective results of system performance derived from these metrics are reported in Section~\ref{sec:Quantitative_Outcomes}.

\subsubsection{Subjective User Experience Metrics}
\label{Subjective_User_Experience}
Subjective evaluations were collected from both participants with dementia and caregivers. 
Users with dementia completed a short User Experience Questionnaire Plus (UEQ+) survey (Appendix~\ref{Participant_survey}), 
while caregivers filled out a structured evaluation form after each session (Appendix~\ref{Caregiver_survey}). To assess user experience, we used a simplified version of the  (UEQ+)~\cite{schrepp2019design} adapted to a 5-point Likert scale 
(1 = strongly disagree, 5 = strongly agree). This modification was made 
to reduce the burdon on participants, as our target population included older adults and individuals with early-stage dementia. Responses were converted 
to numerical values (1–5), and mean scores were computed for each dimension. 
For reference, we also report an equivalent mapping to the original 7-point 
UEQ+ scale using a linear rescaling:
\begin{equation}
    x' = (x-1)\times \tfrac{6}{4} + 1
\end{equation}
where $x$ is the score on the 5-point scale and $x'$ is the rescaled 7-point value. 

In addition to the UEQ+ scale, participants and caregivers were invited to answer eight open-ended questions (5 open-ended questions for users with dementia (Appendix~\ref{Participant_survey}) and 3 for caregivers (Appendix~\ref{Caregiver_survey})) designed to capture qualitative impressions of the interaction. These questions focused on perceived emotional engagement, comfort during dialogue, and suggestions for improving the robot’s behavior and responsiveness. \textbf{Participant open-ended questions.}
Participants with dementia responded to five open-ended questions focusing on (i) aspects of the interaction they found enjoyable, (ii) challenges or difficulties encountered, (iii) suggestions for improving the robot’s behavior, and (iv) willingness for long-term use and recommendation. \textbf{Caregiver open-ended questions.}
Caregivers responded to three open-ended questions capturing (i) observed emotional and behavioral responses of the participant, (ii) perceived impact on mood and social engagement, and (iii) suggestions for system improvement.

These  qualitative responses complemented the quantitative UEQ+ dimensions by highlighting user preferences, emotional resonance, and perceived empathy in the robot's behavior. The subjective results derived from these metrics are reported in Section~\ref{sec:User_Experience_and_Caregiver_Evaluation}.

\subsubsection{Automatic Multimodal Interaction Analysis}
\label{subsec:multimodal_analysis}

To complement subjective ratings and reduce reliance on manual coding, we introduce an
automatic multimodal analysis framework that uses LLMs and VLMs to quantitatively evaluate participant engagement during dementia--robot reminiscence interactions.
The proposed framework captures both \emph{semantic verbal engagement}, reflected in
spoken responses, and \emph{nonverbal engagement}, reflected in gaze orientation
and facial affect.

\paragraph{Semantic Engagement (LLM-based)}
We employed an LLM (GPT-4o) to annotate each participant utterance $u$ along four complementary semantic dimensions that characterize engagement
with the reminiscence prompt and the ongoing interaction.

Specifically, the LLM assigned:
(i) \emph{on-topic relevance} $on\_topic(u)\in\{0,1\}$, indicating whether the utterance was semantically related to the robot’s prompt or displayed media;
(ii) \emph{reminiscence depth} $r\_depth(u)\in\{0,1,2,3\}$, reflecting the extent
of autobiographical content, ranging from minimal or factual responses to rich,
multi-event personal narratives;
(iii) \emph{emotional tone} $emo\_tone(u)$, drawn from four categories
\{POSITIVE, BITTERSWEET, NEGATIVE\_DISTRESS, NEUTRAL/MIXED\}; and
(iv) \emph{dialog act} $da(u)$, capturing the functional role of user utterance
(e.g., answering, self-disclosure, repair).

A summary of the LLM-based annotation scheme is provided in
Table~\ref{tab:llm_semantic_scheme}.

\begin{table}[t]
\centering
\caption{LLM-based semantic verbal engagement annotation scheme.}
\label{tab:llm_semantic_scheme}
\begin{tabular}{p{0.97\linewidth}}
\toprule

\textbf{\textit{$on\_topic(u)$}} — On-topic relevance of the utterance:
\begin{itemize}
  \item 0: the utterance content is \emph{not} semantically related to the current reminiscence prompt or displayed media;
  \item 1: the utterance content is semantically related to the prompt or media, including reflective or affective responses.
\end{itemize}
\\[-0.9em]
\midrule

\textbf{\textit{$r\_depth(u)$}} — Reminiscence depth reflecting autobiographical richness:
\begin{itemize}
  \item 0: no personal content (e.g., yes/no or filler responses);
  \item 1: a single personal detail or value;
  \item 2: multiple personal details or reflective elaboration;
  \item 3: rich autobiographical narrative involving connected experiences.
\end{itemize}
\\[-0.9em]
\midrule

\textbf{\textit{$emo\_tone(u)$}} — Emotional tone capturing dementia-relevant affective expression:
\begin{itemize}
  \item POSITIVE: clearly pleasant, appreciative, or joyful affect;
  \item BITTERSWEET: mixed but meaningful affect, such as reminiscence-related sadness co-occurring with appreciation; this category captures non-harmful sadness common in dementia reminiscence;
  \item NEGATIVE\_DISTRESS: clear emotional distress, discomfort, or expressed desire to avoid the topic;
  \item NEUTRAL/MIXED: primarily factual, ambiguous, or emotionally flat responses without a dominant tone.
\end{itemize}
\\[-0.9em]
\midrule

\textbf{\textit{$da(u)$}} — Dialog act category describing the functional role of the utterance in interaction:
\begin{itemize}
  \item ANSWER: brief responses directly answering the robot’s question;
  \item SELF\_DISCLOSURE: autobiographical sharing beyond minimal yes/no responses;
  \item QUESTION: utterances seeking information or clarification;
  \item ACKNOWLEDGMENT/BACKCHANNEL: brief signals indicating attention or listening without adding new content;
  \item PHATIC/SOCIAL: social remarks not directly related to reminiscence content;
  \item OFF\_TOPIC\_COMMENT: content clearly unrelated to the prompt or displayed media;
  \item REPAIR/MISUNDERSTANDING: utterances indicating hearing, comprehension, or interactional difficulties.
\end{itemize}
\\[-0.9em]

\bottomrule
\end{tabular}
\end{table}

\paragraph{Nonverbal Engagement (VLM-based)}

While verbal content provides critical insight into cognitive and emotional engagement, nonverbal behavior offers complementary signals that are especially important in dementia-related interactions. To capture such cues in a scalable and robust manner, we analyze user-facing video clips corresponding to each dialog turn using a VLM-based nonverbal annotation pipeline.

For each user turn $u$, we sample frames from the associated video segment at 1~Hz and annotate each frame with two coarse nonverbal signals:  
(i) attention toward the robot or tablet interaction area, and  
(ii) facial valence, labeled as negative ($-1$), neutral ($0$), or positive ($+1$).

Frame-level annotations are then aggregated into turn-level nonverbal engagement metrics, including the proportion of frames indicating attention toward the robot or tablet ($gaze\_robot\_ratio(u)$) and summary statistics of facial valence. This aggregation strategy aligns nonverbal engagement metrics with the LLM-based semantic verbal engagement annotations while reducing sensitivity to momentary noise at the frame level.

The resulting nonverbal annotation scheme and derived metrics are summarized in Table~\ref{tab:cv_nonverbal_scheme}.

\begin{table}[t]
\centering
\caption{VLM-based nonverbal engagement annotation scheme (turn-level).}
\label{tab:cv_nonverbal_scheme}
\begin{tabular}{p{0.97\linewidth}}
\toprule

\textbf{\textit{$gaze\_robot\_ratio(u)$}} — Proportion of time the participant attends to the robot or tablet during the turn:
\begin{itemize}\itemsep0pt
  \item Computed as the fraction of sampled frames labeled with attention toward the robot or tablet interaction area.
  \item Captures sustained visual or postural engagement rather than eye contact with the camera.
\end{itemize}
\\[-0.9em]
\midrule

\textbf{\textit{$valence\_mean(u)$}} — Average facial valence during the turn:
\begin{itemize}\itemsep0pt
  \item Frame-level facial valence is labeled as negative ($-1$), neutral ($0$), or positive ($+1$).
\end{itemize}
\\[-0.9em]
\midrule

\textbf{\textit{$valence\_pos\_ratio(u)$}} — Proportion of positively valenced frames:
\begin{itemize}\itemsep0pt
  \item Measures the prevalence of positive facial expressions during the turn.
\end{itemize}
\\[-0.9em]

\bottomrule
\end{tabular}
\end{table}

\section{Results}
\label{sec:results}

\begin{table}[t]
  \centering
  \caption{Session Summary per Participant (N=12). Five additional older-adult sessions were conducted but excluded from analysis due to missing signed consent documentation required by the IRB-approved protocol; engineering/debug sessions are not included.}
  \label{tab:session_summary}
  \begin{tabularx}{\linewidth}{
    >{\centering\arraybackslash}p{0.4cm}   
    >{\centering\arraybackslash}p{0.4cm}   
    >{\centering\arraybackslash}X
    >{\centering\arraybackslash}X
    >{\centering\arraybackslash}X
    >{\centering\arraybackslash}X
    >{\centering\arraybackslash}X
    >{\centering\arraybackslash}X}
    \toprule
    ID & Sex & Age Range & Group & Robot Setup & Duration (min) & Dialogue Rounds & Triggers Used \\
    \midrule
    P04 & F & 75–84 & Healthy & I & 21 & 20 & 4 \\
    P06 & F & 75–84 & PLWD & I & 23 & 17 & 3 \\
    P07 & M & 75–84 & PLWD & I & 24 & 15 & 3 \\
    P08 & F & 85+ & PLWD & I & 17 & 15 & 2 \\
    P09 & M & 85+ & PLWD & I & 22 & 26 & 3 \\
    P10 & F & 85+ & PLWD & I & 5 & 4 & 1 \\
    P11 & F & 75–84 & PLWD & I & 16 & 17 & 2 \\
    P12 & F & 75–84 & PLWD & II & 30 & 17 & 3 \\
    P13 & M & 75–84 & PLWD & II & 15 & 24 & 4 \\
    P14 & M & 85+ & PLWD & II & 10 & 4 & 2 \\
    P15 & F & 65–74 & PLWD & II & 10 & 3 & 2 \\
    P16 & F & 85+ & PLWD & II & 29 & 34 & 4 \\
    \bottomrule
  \end{tabularx}
\end{table}

\subsection{Quantitative System Performance}
\label{sec:Quantitative_Outcomes}

\begin{table}[t]
\caption{Summary of Quantitative System Evaluation Results}
\label{tab:quant_results}
\centering
\begin{tabular}{lcc}
\toprule
\textbf{Metric} & \textbf{Pepper-native} & \textbf{Jetson-edge} \\
\midrule
End-to-end latency (s) & $9.25 \pm 2.81$ & $\textbf{5.89} \pm \textbf{1.11}$ \\
Interaction completion rate & 84.82\% & 100\% \\
Dialogue coherence rate &  88.30\% & 92.28\% \\
\midrule
\textbf{Session-level statistics} & & \\
Dialogue turns per session & \multicolumn{2}{c}{$16.81 \pm 9.27$} \\
Session duration (min) & \multicolumn{2}{c}{$17.94 \pm 7.15$} \\
Participants switching triggers & \multicolumn{2}{c}{13 (mean = 2.44, SD = 1.01)} \\
\bottomrule
\end{tabular}
\end{table}

\subsubsection{Latency and Responsiveness Benchmark}

The study comprised sixteen on-site sessions with sixteen unique participants (P01–P16). Their ages ranged from 50 to 95 years. The demographic distribution of participants, including age range, gender composition, session lengths, and number of dialogue turns per participant, is summarized in Table~\ref{tab:session_summary}. 
Sessions P01–P05 involved healthy older adults interacting through Pepper’s native hardware, while
P06–P11 involved persons living with dementia (PLWD) using the same configuration (\textit{Robot I}). In contrast,
sessions P12–P16 used \textit{Robot II}, a Jetson-based edge client with PLWD.
This design enables a direct comparison between the two human cohorts and the different hardware architectures.

To assess the deployability and performance advantages of the proposed hybrid edge–cloud framework,
we compared two identical system configurations: one using the Pepper robot’s native hardware and the other using 
an external Jetson-based edge client. This comparison aimed to quantify improvements
in real-time responsiveness, modular scalability, and overall robustness achieved by decoupling multimodal processing from the robot’s onboard control hardware. This evaluation serves to
validate the practicality of the proposed architecture for scalable, low-latency deployment in scenarios involving embodied social interaction. 
The average end-to-end latency (from the end of speech to the onset of robot playback)
was $9.25\pm2.81$\,s for the Pepper-native configuration and $5.89\pm1.11$\,s for the Jetson edge client, representing 
a \textbf{36.3\%} reduction in latency (Table~\ref{tab:quant_results}), along with improvements in interaction completion and dialogue coherence. To better understand the source of this improvement, we further decompose the end-to-end latency into component-level contributions, as summarized in Table~\ref{tab:latency_breakdown}. The results indicate that multimodal reasoning contributes a comparable portion of latency in both configurations, as inference is executed on the same host system. In contrast, the Pepper-native setup incurs substantially higher system-level orchestration overhead due to less efficient data transfer, synchronization, and control flow.

\begin{table}[t]
\centering
\caption{Component-Level Latency Breakdown Across System Configurations}
\label{tab:latency_breakdown}
\begin{tabular}{lcc}
\toprule
\textbf{Component} & \textbf{Pepper-native (s)} & \textbf{Jetson-edge (s)} \\
\midrule
Speech recognition (ASR) & 1.18 & 0.97 \\
Multimodal reasoning & 4.63 & 4.48 \\
System orchestration & 3.44 & 0.44 \\
\midrule
\textbf{Total} & 9.25 & 5.89 \\
\bottomrule
\end{tabular}
\end{table}

\subsubsection{Robot Dialogue Coherence and Stability}

Two independent raters judged $92.28\%$ of all responses from Robot II as contextually coherent with prior dialogue turns, compared to $88.30\%$ for Robot I. The interaction completion rate—defined as the proportion of interactions that occurred without technical interruptions or manual intervention from researchers—was 84.82\% for Robot I and 100\% for Robot II, suggesting greater runtime robustness in Robot II.

On average, each session contained $16.81\pm9.27$ dialogue turns and lasted for $17.94\pm7.15$ minutes.  
11 participants switched between multiple memory triggers (mean = 2.44, SD = 1.01).  
System-level logging confirmed continuous multimodal synchronization, as there was nopacket loss or missing frames, which was verified by manual comparision with GoPro recordings.

\subsection{User Experience and Caregiver Evaluation}
\label{sec:User_Experience_and_Caregiver_Evaluation}

\begin{table}[t]
  \centering
  \caption{Participant (dementia) evaluation scores (Likert 1--5) aggregated by interaction dimension. Each dimension groups related survey items (Q1--Q7) corresponding to perceived attractiveness, interaction quality, and usability of the robot system. Values are reported as mean $\pm$ standard deviation across all sessions.}
  \label{tab:caregiver_likert}
  \begin{tabularx}{0.7\linewidth}{l c c}
    \toprule
    \textbf{Dimension} & \textbf{Items} & \textbf{Mean $\pm$ SD} \\
    \midrule
    Attractiveness & Q1–Q2 & 4.7 $\pm$ 0.5 \\
    Interaction quality & Q3–Q5 & 4.5 $\pm$ 0.7 \\
    Usability & Q6-Q7 & 4.0 $\pm$ 0.9 \\
    \bottomrule
  \end{tabularx}
\end{table}

\begin{table}[t]
  \centering
  \caption{Caregiver ratings by dimension (Likert 1–5). Each dimension groups related survey items (Q1--Q9) corresponding to interface usability, emotional engagement, dialogue naturalness, system stability, and overall recommendation.}
  \label{tab:caregiver_likert}
  \begin{tabularx}{0.9\linewidth}{l c c}
    \toprule
    \textbf{Dimension} & \textbf{Items} & \textbf{Mean $\pm$ SD} \\
    \midrule
    Interface usability & Q1–Q2 & 4.6 $\pm$ 0.7 \\
    Emotional/positive engagement & Q3, Q8 & 3.7 $\pm$ 1.5 \\
    Linguistic/dialogue naturalness & Q4–Q6 & 3.5 $\pm$ 1.2 \\
    System stability & Q7 & 3.7 $\pm$ 1.0 \\
    Overall recommendation & Q9 & 4.5 $\pm$ 0.6 \\
    \bottomrule
  \end{tabularx}
\end{table}

\subsubsection{Quantitative Survey Outcomes}
\label{sec:Quantitative_Survey_Outcomes}

The quantitative survey results show consistently positive evaluations from both participants with dementia and their caregivers. 
Participants rated the system highly across three assessed UEQ+ dimensions, with mean scores of 
$4.7 \pm 0.5$ for \textit{attractiveness}, $4.5 \pm 0.7$ for \textit{interaction quality}, and $4.0 \pm 0.9$ for \textit{usability}. 
The internal consistency of the UEQ+ items was acceptable ($\alpha = 0.82$), suggesting reliable measurement of subjective experience. 

Caregiver assessments used a 5-point Likert scale (1 = strongly disagree, 5 = strongly agree) across nine items (Q1–Q9). 
To enhance interpretability, the items were grouped into five dimensions: 
\textit{Interface usability} (Q1–Q2), \textit{Emotional/positive engagement} (Q3, Q8), 
\textit{Linguistic/dialogue naturalness} (Q4–Q6), \textit{System stability} (Q7), 
and \textit{Overall recommendation} (Q9). 
Mean caregiver ratings across these dimensions were consistently above 3.5. The highest evaluations were for 
interface usability ($4.6 \pm 0.7$) and system stability ($3.7 \pm 1.0$), 
reflecting intuitive interaction and technical reliability in practice. 
Emotional engagement ($3.7 \pm 1.5$) and dialogue naturalness ($3.5 \pm 1.2$) were also rated positively, 
suggesting that the robot’s adaptive responses were perceived as supportive and contextually appropriate. 
The overall recommendation score reached $4.5 \pm 0.6$, 
indicating strong endorsement for continued use in reminiscence activities. This also indicates that the proposed edge architecture enhances technical efficiency without compromising user-perceived quality or engagement.

\subsubsection{Qualitative Observations and Thematic Insights}

In addition to the quantitative outcomes reported in Section~\ref{sec:Quantitative_Survey_Outcomes}, we conducted semi-structured interviews with both users with dementia and caregivers to capture qualitative perspectives on usability, emotional engagement, and long-term acceptance (See Section~\ref{Subjective_User_Experience}). 

\paragraph*{User Perspectives.}
Across both configurations, users reported that the reminiscence interactions were intuitive, engaging, and emotionally positive. Many expressed enjoyment in discussing personal memories with the robot, noting that the robot’s ability to recall and display familiar images made conversations feel meaningful and personalized. Participants interacting through the Jetson-based system particularly emphasized its  smoother response timing, clearer voice output, and sharper screen display, which enhanced the perceived naturalness of the interaction and reduced confusion. Minor challenges included initial hesitation with using buttons and limited speech recognition accuracy in noisy environments; however, these issues diminished with familiarization. Most participants expressed a willingness to engage in repeated sessions and recommended the system for continued use in memory-support programs.

\paragraph*{Caregiver Perspectives.}
Caregivers observed that users appeared more alert and emotionally expressive when interacting with the robot compared to typical daily conversations. Several caregivers reported that the system promoted smoother turn-taking and increased sustained attention, particularly when the robot adapted its tone or the length of prompts based on user responses. Notably, caregivers in the Jetson condition highlighted improvements in dialogue fluidity and audiovisual quality, describing the robot’s speech and display as more “lifelike” and “easy to follow.” They also praised the simplified interaction interface and the stable system operation, citing fewer delays and interruptions during sessions.
Constructive suggestions included expanding the repertoire of conversation themes, adding multimodal prompt cues (e.g., online videos or gestures), and enabling caregivers with more flexible control through the web portal.

\paragraph*{Thematic Insights.}
Responses from both patients and caregivers were thematically coded, revealing three dominant patterns that align with the design goals of the framework, which aims to explain \textit{why} the proposed edge–cloud, caregiver-in-the-loop architecture improved the perceived quality and user experience of human–robot interaction: (1) Affective resonance: Participants consistently described interactions as emotionally meaningful and personally engaging.  
This theme reflects how personalized media and responsive dialogue fostered empathy and memory recall—validating the framework’s ability to support emotion-aware, personalized reminiscence. (2) Interaction fluidity: Both user groups emphasized the importance of system speed, timing, and audiovisual clarity.  
Sessions conducted with the Jetson-based edge client were frequently characterized as smoother and more natural, supporting our technical hypothesis that low-latency local inference enhances conversational flow and engagement continuity. (3) Personalization and adaptability: Users and caregivers highlighted the value of an agent that ``learns” from previous sessions and adapts to individual preferences.  
This theme reinforces the need for cross-session memory and adaptive prompt generation—key capabilities enabled by the distributed architecture and MemoryLane integration.

Together, these insights demonstrate that the technical enhancements in the proposed system—particularly edge-level inference and multimodal personalization—directly translate into measurable improvements in the quality of social and emotional patient-robot interactions.  

\subsection{Automatic Multimodal Interaction Results}

Using the proposed LLM- and VLM-based annotation pipeline, we computed turn-level verbal and nonverbal engagement metrics across all sessions. These metrics enable quantitative analysis of both what participants expressed verbally and how they engaged nonverbally during interaction.

\subsubsection{Validation with Human Coders}

To assess the reliability of the proposed automatic annotations, we conducted a limited validation study with trained human coders. A randomly selected subset of sessions was manually annotated by two independent coders following the same annotation definitions described in Table~\ref{tab:llm_semantic_scheme} and ~\ref{tab:cv_nonverbal_scheme}. 

For verbal content, coders labeled on-topic relevance, reminiscence depth, emotional tone, and dialog act category at the utterance level. For nonverbal behavior, coders annotated attention toward the robot or tablet interaction area and coarse facial valence on sampled video frames. Table~\ref{tab:case_p08_turnlevel_valence} shows an example of the annotation for P08. Automatic LLM-based and VLM-based annotations were then compared against human labels using percentage agreement. Rather than treating human annotations as ground truth, we interpret these agreement measures as indicators of consistency between automatic and human judgments. Given the known challenges of speech recognition, affect interpretation, and visual analysis in dementia populations, the goal of this validation is not to establish clinical accuracy, but to demonstrate that the proposed automatic metrics capture interaction patterns that are broadly aligned with human interpretation.

\begin{table}[t]
\centering
\caption{Illustrative turn-level multimodal engagement for Participant P08.}
\label{tab:case_p08_turnlevel_valence}
\setlength{\tabcolsep}{4pt}
\renewcommand{\arraystretch}{1.1}

\begin{tabular}{c c c c c c c c}
\toprule
\textbf{Turn} &
$t_{\text{user}}$ (s) &
$on\_topic$ &
$r\_d$ &
$emo$ &
$da$ &
$gaze\_r$ &
$val+$ \\

\midrule
1  & 10 & 1 & 1 & POS  & DIS & 1 & 0.667 \\
2  & 8 & 1 & 1 & BIT  & DIS & 1 & 0.875 \\
3  & 30 & 1 & 2 & BIT  & DIS & 0.967 & 0.9 \\
4  & 39 & 1 & 2 & POS  & DIS & 1 & 0.842 \\
5  & 54 & 1 & 2 & POS  & DIS & 1 & 0.226 \\
6  & 41 & 1 & 2 & BIT  & DIS & 1 & 0.25 \\
7  & 45 & 1 & 2 & POS  & DIS & 1 & 0.477 \\
8  & 12 & 1 & 1 & NEU  & DIS & 1 & 0 \\
9  & 29 & 1 & 2 & POS  & DIS & 1 & 0.143 \\
10  & 40 & 1 & 2 & POS  & DIS & 1 & 0.077 \\
11  & 9 & 1 & 1 & NEU  & DIS & 1 & 0.5 \\
12  & 37 & 1 & 2 & POS  & DIS & 1 & 0.889 \\
13 & 36 & 1 & 2 & BIT  & DIS & 0.971 & 0.771 \\
14 & 21 & 0 & 0 & NEU  & QUE & 1 & 0.35 \\
\bottomrule
\end{tabular}

\vspace{2pt}
\footnotesize
\textit{Notes:} $t_{\text{user}}$ (s) indictae the User speaking time in this turn. Abbreviations are defined in Table~\ref{tab:llm_semantic_scheme} and ~\ref{tab:cv_nonverbal_scheme}.
\end{table}

\subsubsection{Session-Level Engagement Statistics}
We first summarize engagement metrics at the session level by aggregating turn-level annotations. Table~\ref{tab:participant_engagement} reports mean and standard deviation of key semantic and nonverbal metrics for each robot condition. Across participants, the system elicited consistently high on-topic response rates and moderate-to-high levels of autobiographical depth and self-disclosure, indicating that the robot was generally successful in sustaining semantically relevant and personally meaningful dialogue. Importantly, nonverbal engagement, as measured by gaze toward the robot or tablet, remained high across most participants, even in cases where verbal contribution was limited.

\begin{table}[t]
\centering
\caption{Participant-level multimodal engagement statistics aggregated across dialog turns.}
\label{tab:participant_engagement}
\begin{tabular}{l >{\centering\arraybackslash}m{0.6cm} cccccc}
\toprule
ID &
Robot Setup &
On-topic &
$r_d$ &
SDis &
Gaze &
Val-mean &
Val+ \\
\midrule
P06 & I & 0.92 & 1.58 & 0.92 & 0.94 & $-$0.23 & 0.00 \\
P07 & I & 0.83 & 1.33 & 0.83 & 0.45 & $-$0.47 & 0.00 \\
P08 & I & 0.93 & 1.57 & 0.93 & 1.00 &  0.30 & 0.50 \\
P09 & I & 0.64 & 0.80 & 0.52 & 0.78 & $-$0.12 & 0.00 \\
P10 & I & 0.00 & 0.00 & 0.00 & 1.00 &  0.00 & 0.00 \\
P11 & I & 0.50 & 0.61 & 0.44 & 1.00 &  0.64 & 0.61 \\
P12 & II & 0.50 & 0.83 & 0.44 & 0.92 &  0.02 & 0.03 \\
P13 & II & 0.76 & 0.71 & 0.48 & 0.99 &  0.05 & 0.04 \\
P14 & II & 0.50 & 0.50 & 0.25 & 1.00 &  0.36 & 0.31 \\
P15 & II & 0.33 & 0.22 & 0.22 & 1.00 &  0.92 & 0.89 \\
P16 & II & 0.66 & 0.75 & 0.44 & 1.00 & $-$0.11 & 0.02 \\
\midrule
Mean & & 0.60 & 0.81 & 0.50 & 0.92 &  0.12 & 0.22 \\
SD &  & 0.27 & 0.51 & 0.29 & 0.17 &  0.40 & 0.31 \\
\bottomrule
\end{tabular}

\vspace{2pt}
\footnotesize{
\textit{Notes:} 
\textbf{SDis}: proportion of turns labeled as SELF\_DISCLOSURE;
\textbf{Gaze}: mean gaze\_robot\_ratio across turns;
\textbf{Val+}: mean proportion of positively valenced frames.
}
\end{table}

\subsubsection{Relationship Between Verbal and Nonverbal Engagement}
To examine the coupling between verbal and nonverbal engagement, we analyzed correlations between LLM-based semantic metrics and VLM-based attention measures. As shown in Table~\ref{tab:participant_engagement}, the dissociation between verbal and nonverbal engagement is particularly salient in dementia-related interactions. Some participants produced relatively verbose speech while showing limited or inconsistent visual or postural attention, a pattern that would similarly be mischaracterized as high engagement by language-only metrics and is also consistent with known behavioral presentations in dementia~\cite{jiang2024comprehension, badal2024investigating}. Conversely, few participants exhibited sustained visual or postural attention despite producing fewer or shorter utterances, a pattern that would be underestimated by language-only metrics. By integrating LLM-based semantic analysis with VLM-based nonverbal cues, our framework captures a more complete and ecologically valid picture of engagement than unimodal approaches. Furthermore, From a practical perspective, these metrics enable longitudinal, fine-grained monitoring of engagement trajectories without requiring extensive human coding. Such automated multimodal summaries could support caregivers and clinicians in identifying changes in interaction quality over time, detecting disengagement or distress, and tailoring intervention strategies accordingly. 

\subsubsection{Comparison Across Robot Setups}

To examine whether different interaction configurations lead to systematic differences in engagement, we further compared multimodal engagement metrics across two robot setups. Participants P06--P11 interacted with Setup~I, while participants P12--P16 interacted with Setup~II. For each participant, engagement metrics were first aggregated across dialog turns, and group-level statistics were then computed within each setup.

Table~\ref{tab:setup_comparison} summarizes the mean and standard deviation of key semantic and nonverbal engagement metrics across the two robot setups. Rather than uniformly outperforming one another, the two setups exhibit complementary engagement profiles. Setup~I is associated with higher on-topic response rates, greater reminiscence depth, and a higher proportion of self-disclosure, indicating stronger semantic coherence and richer autobiographical engagement in user responses. In contrast, Setup~II demonstrates higher gaze-to-robot ratios and more positive facial valence, suggesting enhanced nonverbal engagement and affective involvement during interaction.

While this study is not powered for formal statistical hypothesis testing, the consistent trends observed across multiple independent metrics suggest that interaction configuration can meaningfully influence both verbal and nonverbal engagement patterns. These results highlight the sensitivity of the proposed LLM- and VLM-based evaluation framework to design-level differences in real-world dementia–robot interactions. More importantly, the proposed LLM–VLM-based evaluation framework enables systematic, longitudinal tracking of both semantic and nonverbal engagement patterns. This capability is particularly valuable for future long-term deployments, where caregivers and clinicians may wish to monitor gradual changes in interaction quality, affective expression, and attentional engagement over time.

\begin{table}[t]
\centering
\caption{Comparison of multimodal engagement metrics across robot setups.}
\label{tab:setup_comparison}
\begin{tabular}{lcc}
\toprule
Metric & Robot Setup~I & Robot Setup~II \\
\midrule
On-topic rate              & 0.64 $\pm$ 0.35 & 0.55 $\pm$ 0.16 \\
Mean reminiscence depth    & 0.98 $\pm$ 0.63 & 0.60 $\pm$ 0.25 \\
Self-disclosure ratio      & 0.61 $\pm$ 0.36 & 0.37 $\pm$ 0.12 \\
Gaze-to-robot rate         & 0.86 $\pm$ 0.22 & 0.98 $\pm$ 0.04 \\
Valence mean               & 0.02 $\pm$ 0.39 & 0.25 $\pm$ 0.41 \\
Positive valence ratio     & 0.18 $\pm$ 0.29 & 0.26 $\pm$ 0.37 \\
\bottomrule
\end{tabular}
\end{table}

\section{Discussion}

\subsection{System-Level Implications for Edge-host-cloud SAR in Dementia Care}
Our findings demonstrate that the proposed hybrid edge–cloud framework effectively enables emotion-aware and personalized reminiscence interactions in real-world dementia-care settings. Compared to prior socially assistive robotics systems that rely on fully onboard processing or tightly coupled architectures~\cite{hsieh2023socially,liao2023use}, our distributed edge–host design improves scalability and responsiveness while preserving embodied interaction. In contrast to pre-scripted or rule-based dialogue systems commonly used in dementia-care SAR~\cite{yuan2021systematic}, the integration of multimodal reasoning and adaptive dialogue generation enables more flexible and context-sensitive interactions. Both participants and caregivers consistently reported positive user experience, while the quantitative system evaluations confirmed the framework’s robustness and technical feasibility across heterogeneous hardware setups. Importantly, the results highlight a bidirectional relationship between system responsiveness and affective engagement: the reduced latency and improved output consistency achieved by the Jetson edge client supported smoother conversational rhythm and more coherent dialogue flow, thereby aligning low-level system optimization with high-level social interaction quality. Beyond validating performance on different Robot Setup, the modular design of our framework—decoupling perception, reasoning, and actuation—supports architectural generalizability. Because all multimodal processing modules follow standardized communication interfaces, the same pipeline can be redeployed on other socially assistive robots or embedded platforms with minimal adaptation. This provides a reusable foundation for future SAR systems requiring scalable multimodal processing. Furthermore, our automatic LLM-based semantic and CV-based nonverbal engagement metrics offer a replicable methodology for evaluating SAR–dementia interactions. These quantitative indicators (e.g., reminiscence depth, on-topic coherence, affective valence, gaze-to-robot behavior) complement traditional caregiver-reported outcomes~\cite{yuan2021systematic,otake2021cognitive} and allow fine-grained, objective monitoring of how technical system properties—such as latency, output timing, and multimodal synchronization—shape patient engagement. Together, these insights contribute toward establishing a principled evaluation framework for future emotion-aware SAR systems in dementia care.

\subsection{Role of Caregiver-in-the-Loop Personalization in Dementia-Care Robotics}
From a clinical and caregiving perspective, the reduction in response latency and improvement in dialogue stability are not only technical achievements; they also significantly influence user engagement and comfort. In dementia care, conversational timing is crucial for attention and comprehension; long pauses can interrupt recall or cause anxiety~\cite{yuan2021systematic,abdollahi2022artificial}. The observed 36$\%$ reduction in latency, therefore, has practical relevance for sustaining focus and minimizing frustration during reminiscence sessions. Caregivers reported that sessions flowed more naturally and were easier to facilitate compared to typical reminiscence conversations without the robot. This reflects the important caregiver-in-the-loop personalization role: caregivers continuously monitor residents’ emotional states, attentional shifts, and fatigue levels, and the robot’s improved responsiveness helped them maintain an appropriate conversational rhythm with less manual intervention. Consistent pacing and automated turn-taking may reduce caregiver workload and cognitive load while preserving resident participation—outcomes that align closely with person-centered goals in long-term dementia care. The integration of personalized photographs and memory prompts further aligns the system with established reminiscence-therapy principles. By combining individualized media with emotion-aware dialogue, the framework effectively operationalizes person-centered communication in a scalable, technology-enabled form. Such adaptive interaction may complement existing therapeutic and palliative-care approaches, potentially extending the reach of structured engagement to settings with limited staffing. Future clinical research should evaluate whether these immediate engagement benefits translate into measurable outcomes, such as improved mood, reduced agitation, or enhanced caregiver satisfaction over repeated sessions. Longitudinal evaluations in routine care environments will be essential to determine the durability of benefits and the feasibility of workflow integration.

At a broader level, our findings suggest that distributed multimodal frameworks can bridge the gap between technical performance and human emotional experiences in socially assistive robotics. This paradigm—incorporating local perception and adaptation while maintaining global coordination—may extend beyond reminiscence care and apply to other forms of personalized social interaction.

\subsection{Lessons Learned for Dementia–Robot Interaction}
1) Inter-individual variability and the need for adaptive perception: In our deployment, we observed substantial inter-individual variability in how people with dementia engaged with the robot. For example, two participants (P10 and P15) completed only 2–4 brief interaction rounds before losing interest, whereas others eagerly engaged across sessions, voluntarily shared rich personal memories and emotions with the robot, and explicitly expressed appreciation when it displayed their family photos. Participant P10 even showed signs of resistance and agitation before the session began. These contrasting patterns highlight a key challenge for dementia-focused HRI: even when the robot’s capabilities and scripted task remain identical, a person’s moment-to-moment willingness and ability to interact can vary greatly.

This suggests that scaling such systems will require more than fixed dialogue flows; it will require automated perception and adaptive policies capable of inferring (or at least approximating) users’ current cognitive state, affect, and willingness to engage, as well as awareness of other agents and activities in the environment. Technically, this motivates future work on multimodal engagement estimation, online policy adaptation, and interaction strategies that can gracefully back off, re-engage, or hand over to human caregivers in response to subtle behavioral cues.

2) Contextual and infrastructural constraints on real-world deployment:
Our study also underscored that deploying social robots in real-world dementia care settings is constrained not only by robot and user factors, but also by contextual infrastructure. Across all three study phases, the robot could not reliably connect to the existing institutional Wi-Fi networks, requiring the research team to bring and configure a dedicated router to enable the system to function. This illustrates that strong performance in controlled lab environments does not straightforwardly translate to robustness in real-world settings, where network policies, security constraints, and physical layout may differ substantially. For social robots to become a sustainable part of care ecosystems, future work must address these implementation-level challenges—for example, through designs that are less dependent on institutional networks (e.g., local edge computation and offline modes), standardized deployment procedures, and closer integration with IT and clinical workflows. More generally, our findings point to the importance of implementation science for robotics: systematically studying how technical, organizational, and environmental factors interact to enable or hinder successful, long-term use of socially assistive robots in dementia care.

\subsection{Lessons Learned for Stakeholder-in-the-Loop Design in Dementia Care}

A central lesson from our field deployment is that successful dementia-care robotics requires not only patient-facing personalization but also stakeholder-in-the-loop design that continuously incorporates the expertise, constraints, and situated practices of caregivers and clinical staff. While the robot provided emotion-aware and media-driven reminiscence support, the overall interaction quality depended heavily on how caregivers prepared participants, interpreted behavioral cues, and intervened when necessary. Their domain knowledge—particularly regarding residents’ cognitive fluctuations, trauma histories, and daily routines—played a critical role in shaping when and how the robot should initiate, pause, or redirect the conversation. This underscores that automated models of engagement must ultimately be grounded in caregiver knowledge rather than treated as purely technical inference problems, consistent with participatory and human-centered design principles widely adopted in socially assistive robotics and healthcare systems~\cite{dautenhahn2009kaspar,hsieh2023socially}. Caregivers also identified opportunities for micro-personalization that fall outside the robot’s current autonomy. For instance, some residents required slower pacing or longer pauses after each prompt, whereas others benefitted from more dynamic conversational scaffolding; some engaged more easily when the robot referenced specific family members. These observations suggest that future systems should support caregiver-configurable “interaction profiles’’—lightweight, adjustable parameters that allow staff to tailor session flow, emotional tone, and prompt selection without reprogramming or technical burden.

Beyond interaction mechanics, stakeholder involvement was also essential for workflow integration. Caregivers emphasized the importance of protocol setup procedures, minimal technical overhead, and compatibility with the rhythms of daily care. Their feedback prompted refinements to session length, trigger selection, and recovery strategies, revealing that the robot’s value lies not only in its conversational capabilities but also in how seamlessly it fits into the existing care ecosystem. Incorporating these insights into system development highlights the need for participatory, iterative design practices that position caregivers as continuous collaborators rather than peripheral users.

Collectively, these lessons reinforce that socially assistive robots for dementia care must be co-designed with the stakeholders who shape their real-world use. Caregiver-in-the-loop mechanisms—ranging from adaptive personalization settings to shared decision-making about engagement strategies—represent a crucial pathway for ensuring that technical innovations translate into meaningful, sustainable practice in long-term care environments.

\subsection{Limitations and Future Work}

Several challenges emerged during the study. First, some participants required caregivers’ assistance to initiate or navigate the interface, highlighting the need for more intuitive interaction modalities such as gesture-based control, simplified touch interfaces, or fully voice-driven operation. Second, intermittent technical interruptions occasionally disrupted the conversational flow, underscoring the importance of fault-tolerant buffering, automatic reconnection, and recovery strategies for real-world deployment. Third, although GPT-4o enabled flexible and coherent dialogue, its outputs occasionally lacked cultural sensitivity, dementia-appropriate language simplification, or personalized grounding. Addressing these issues will likely require domain-specific fine-tuning, controlled decoding strategies, and integration of structured knowledge sources. 

A related limitation is the system’s dependence on a commercial cloud-based large language model. While this provided high-quality language generation, it also introduced dependencies on external infrastructure, evolving model behavior, and network stability. Future work will explore deploying lighter-weight, locally hosted LLMs and VLMs on the Jetson platform or institutional servers to enhance reproducibility, reduce network dependency, and ensure stronger privacy protection. Establishing a standardized local inference pipeline may also improve robustness and consistency across facilities. The study’s scope was constrained by a modest sample size and single-session design, limiting our ability to assess long-term behavioral changes or adaptation dynamics. Future longitudinal trials—conducted over weeks or months—will be essential for understanding how engagement patterns evolve, whether personalization effects accumulate, and how the robot fits into daily life in dementia-care environments. Continuous, in-the-wild deployments in both assisted-living and home settings will further strengthen ecological validity.

Finally, expanding the MemoryLane portal to support richer multimodal stimuli (e.g., music, family videos, historical objects) and integrating caregiver-in-the-loop preference updates could deepen personalization. Iterative co-design with patients, caregivers, and clinical staff will remain crucial for ensuring cultural sensitivity, usability across cognitive levels, and compatibility with existing care workflows.

\section{Conclusion}
In this work, we presented an embodied robotic implementation of the \textit{Speaking Memories} framework, which integrates a distributed multimodal architecture with implicit affect inference for emotion-aware reminiscence interaction.  This integration improves both user experience and the quality of interactions.   
The proposed system unifies a secure cloud portal for data management, a local interaction server for multimodal inference, and a scalable edge computing client that can be embedded into any robot platform.  
This hybrid design - combining cloud, edge, and embodied components - enables real-time, privacy-preserving dialogue generation while maintaining modularity for cross-platform deployment.
Experimental evaluations conducted on both Jetson-based and Pepper-native configurations demonstrated the framework’s technical feasibility and robustness.  
Quantitative analyses confirmed low latency, coherent dialogue generation, and stable multimodal synchronization. Moreover, qualitative feedback from patients, caregivers, and clinical collaborators highlighted improved engagement, emotional resonance, and usability in real-world dementia care contexts.  
The distributed architecture significantly enhances computational efficiency, scalability, and plug-and-play adaptability, which helps to reduce operational costs and simplifies  deployment across heterogeneous robotic systems.
This work thus contributes a generalizable blueprint for SARs, bridging the gap between engineering performance and affective human–robot interaction.  
Future research will extend this framework towards longitudinal deployments, multi-robot coordination, and adaptive personalization mechanisms that learn continuously from user behavior and contextual history.

\section*{Acknowledgments}
The authors would like to thank the faculty collaborators for their valuable guidance.


{\appendices
\section{Participant Experience Survey}
\label{Participant_survey}

\textbf{Study Title:} Robot-assisted Conversation Study for Older Adults

\textbf{Participant ID:} \underline{\hspace{3cm}} \quad

\textbf{Date:} \underline{\hspace{3cm}}

\vspace{0.4cm}
Please complete the following questionnaire to assess the robot’s functions.
The questionnaire presents pairs of opposing robot properties.
Each pair is rated on a scale represented by five boxes.
Check one box to indicate your level of agreement.

\noindent\textbf{Example:}

\begin{center}
\begin{tabular}{|c|*{5}{c|}c|}
\hline
unattractive & $\square$ &{\rlap{$\square$}\hspace{0.1em}\checkmark}& $\square$  & $\square$ & $\square$ & attractive \\
\hline
\end{tabular}
\end{center}

With this assessment, you state that you consider the robot \textbf{rather} \textbf{unattractive} than \textbf{attractive}. 

Under each group, you can also indicate \textbf{how important this aspect is for the overall evaluation of the robot assistant}.
There is no ``right'' or ``wrong'' answer. Only your personal opinion counts.

\subsection*{Section A: Some questions about you}
Q1. What is your age? \\
$\square$ 18--49 \quad
$\square$ 50--64 \quad
$\square$ 65--74 \quad
$\square$ 75--84 \quad
$\square$ Above 85

Q2. What is your gender? \\
$\square$ Female \quad
$\square$ Male \quad
$\square$ Other: \underline{\hspace{3cm}} \quad
$\square$ Prefer not to say

Q3. Have you ever been diagnosed with any of the following cognitive conditions? \\
$\square$ Mild cognitive impairment (MCI) \\\quad
$\square$ Early-stage or mild dementia (with a Clinical Dementia Rating score (CDR) = 1) \\\quad
$\square$ Moderate-stage or moderate dementia (CDR = 2) \\\quad
$\square$ Late-stage dementia (CDR = 3)\\ \quad
$\square$ None of the above

Q4. If you have been diagnosed with dementia, what type of dementia have you been diagnosed with (Select one item that applies to you)? \\
$\square$ Alzheimer’s disease \\\quad
$\square$ Vascular dementia \\\quad
$\square$ Lewy body dementia \\\quad
$\square$ Frontotemporal dementia\\ \quad
$\square$ Parkinson's disease dementia \\ \quad
$\square$ Early-onset dementia \\\quad
$\square$ Mixed dementia \\\quad
$\square$ Other dementia (Please specify) \underline{\hspace{3cm}} \\ \quad
$\square$ Don’t know or remember \\ \quad
$\square$ Not applicable

Q5. Do you have any experience with the following smart devices? (Check all that apply) \\
$\square$ Smartphone \quad
$\square$ Tablet \quad
$\square$ Smartwatch \quad
$\square$ Smart speaker (e.g., Amazon Echo, Google Home) 
$\square$ Smart TV \quad
$\square$ Other (please specify): \underline{\hspace{3cm}} \quad
$\square$ None

\subsection*{Section B: Robot Guidance and Interaction Quality}

\textbf{In my opinion, the Robot is generally:}

\begin{tabularx}{\columnwidth}{|l|*{5}{>{\centering\arraybackslash}p{0.55cm}|}X|}
\hline
Annoying & $\square$ & $\square$ & $\square$ & $\square$ & $\square$ & Enjoyable \\
\hline
Bad & $\square$ & $\square$ & $\square$ & $\square$ & $\square$ & Good \\
\hline
Unpleasant & $\square$ & $\square$ & $\square$ & $\square$ & $\square$ & Pleasant \\
\hline
Unfriendly & $\square$ & $\square$ & $\square$ & $\square$ & $\square$ & Friendly \\
\hline
\end{tabularx}
\begin{center}
\noindent I consider the robot system property described by these terms as: \\
Completely irrelevant 
$\square$ $\square$ $\square$ $\square$ $\square$ 
Very important
\end{center}

\textbf{In my opinion, the surface (Appearance and touchpad functionality) of the product is:}

\begin{tabularx}{\columnwidth}{|l|*{5}{>{\centering\arraybackslash}p{0.1cm}|}X|}
\hline
Unpleasant to touch   & $\square$ & $\square$ & $\square$ & $\square$ & $\square$ & Pleasant to touch  \\
\hline
Rough   & $\square$ & $\square$ & $\square$ & $\square$ & $\square$ & Smooth   \\
\hline
\end{tabularx}
\begin{center}
\noindent I consider the robot system property described by these terms as: \\
Completely irrelevant 
$\square$ $\square$ $\square$ $\square$ $\square$ 
Very important
\end{center}

\textbf{In my opinion, the response behaviour of the Robot is:}

\begin{tabularx}{\columnwidth}{|l|*{5}{>{\centering\arraybackslash}p{0.5cm}|}X|}
\hline
Artificial & $\square$ & $\square$ & $\square$ & $\square$ & $\square$ & Natural \\
\hline
Unpleasant & $\square$ & $\square$ & $\square$ & $\square$ & $\square$ & Pleasant \\
\hline
Boring & $\square$ & $\square$ & $\square$ & $\square$ & $\square$ & Entertaining \\
\hline
\end{tabularx}

\begin{center}
\noindent I consider the robot system property described by these terms as: \\
Completely irrelevant 
$\square$ $\square$ $\square$ $\square$ $\square$ 
Very important
\end{center}

\textbf{In my opinion, the conversation by the robot is:}

\begin{tabularx}{\columnwidth}{|l|*{5}{>{\centering\arraybackslash}p{0.5cm}|}X|}
\hline
Inappropriate  & $\square$ & $\square$ & $\square$ & $\square$ & $\square$ & Suitable  \\
\hline
Unhelpful  & $\square$ & $\square$ & $\square$ & $\square$ & $\square$ & Helpful  \\
\hline
Unintelligent   & $\square$ & $\square$ & $\square$ & $\square$ & $\square$ & Intelligent  \\
\hline
\end{tabularx}

\begin{center}
\noindent I consider the robot system property described by these terms as: \\
Completely irrelevant 
$\square$ $\square$ $\square$ $\square$ $\square$ 
Very important
\end{center}

\textbf{In my opinion the robot has understood my voice input:}

\begin{tabularx}{\columnwidth}{|l|*{5}{>{\centering\arraybackslash}p{0.55cm}|}X|}
\hline
Incorrectly   & $\square$ & $\square$ & $\square$ & $\square$ & $\square$ & Correctly   \\
\hline
Inaccurate   & $\square$ & $\square$ & $\square$ & $\square$ & $\square$ & Accurate    \\
\hline
Confusing   & $\square$ & $\square$ & $\square$ & $\square$ & $\square$ & Clear   \\
\hline
\end{tabularx}

\begin{center}
\noindent I consider the robot system property described by these terms as: \\
Completely irrelevant 
$\square$ $\square$ $\square$ $\square$ $\square$ 
Very important
\end{center}

\textbf{In my opinion, using the Robot is:}

\begin{tabularx}{\columnwidth}{|l|*{5}{>{\centering\arraybackslash}p{0.55cm}|}X|}
\hline
Difficult & $\square$ & $\square$ & $\square$ & $\square$ & $\square$ & Easy \\
\hline
Confusing & $\square$ & $\square$ & $\square$ & $\square$ & $\square$ & Clear \\
\hline
Unhelpful & $\square$ & $\square$ & $\square$ & $\square$ & $\square$ & Helpful \\
\hline
\end{tabularx}

\begin{center}
\noindent I consider the robot system property described by these terms as: \\
Completely irrelevant 
$\square$ $\square$ $\square$ $\square$ $\square$ 
Very important
\end{center}

\textbf{In my opinion, using this robot in social situations makes me feel:}

\begin{tabularx}{\columnwidth}{|l|*{5}{>{\centering\arraybackslash}p{0.35cm}|}X|}
\hline
Inappropriate  & $\square$ & $\square$ & $\square$ & $\square$ & $\square$ & Suitable  \\
\hline
Embarrassed   & $\square$ & $\square$ & $\square$ & $\square$ & $\square$ & Proud   \\
\hline
Uncomfortable   & $\square$ & $\square$ & $\square$ & $\square$ & $\square$ & Comfortable   \\
\hline
\end{tabularx}

\begin{center}
\noindent I consider the robot system property described by these terms as: \\
Completely irrelevant 
$\square$ $\square$ $\square$ $\square$ $\square$ 
Very important
\end{center}

\subsection*{Section C: Open-Ended Feedback}

Q1. What did you like most about the robot interaction? \vspace{0.5cm} \\
\vspace{0.5cm}\rule{\columnwidth}{0.3pt} \\ \rule{\columnwidth}{0.3pt}

Q2. What challenges or difficulties did you face during the interaction?  \vspace{0.5cm} \\
\vspace{0.5cm}\rule{\columnwidth}{0.3pt} \\ \rule{\columnwidth}{0.3pt}

Q3. What improvements would you suggest for the robot’s behavior or design?  \vspace{0.5cm} \\
\vspace{0.5cm}\rule{\columnwidth}{0.3pt} \\ \rule{\columnwidth}{0.3pt}

Q4. Would you like to use it over long term?  \vspace{0.5cm} \\
\vspace{0.5cm}\rule{\columnwidth}{0.3pt} \\ \rule{\columnwidth}{0.3pt}

Q5. Would you like to suggest the robot to others?  \vspace{0.5cm} \\
\vspace{0.5cm}\rule{\columnwidth}{0.3pt}

\section{Caregiver Evaluation Survey}
\label{Caregiver_survey}

\textbf{Participant ID:} \underline{\hspace{3cm}} \quad

\textbf{Date:} \underline{\hspace{3cm}} \quad

\textbf{Relation:} \underline{\hspace{4cm}}

\textbf{Instructions:} 

Please rate each statement on a scale from 

\textbf{1}-(\textbf{Strongly Disagree}) to \textbf{5}-(\textbf{Strongly Agree}).

\begin{tabular}{|c|p{4cm}|c|c|c|c|c|}
\hline
\textbf{ID} & \textbf{Item} & 1 & 2 & 3 & 4 & 5 \\
\hline
Q1 & The Robot interface and operation were clear and intuitive for the loved one (patient). & $\square$ & $\square$ & $\square$ & $\square$ & $\square$  \\
\hline
Q2 & The MemoryLane interface and operation were clear and intuitive for me. & $\square$ & $\square$ & $\square$ & $\square$ & $\square$ \\
\hline
Q3 & The patient displayed positive affect during the interaction. & $\square$ & $\square$ & $\square$ & $\square$ & $\square$  \\
\hline
Q4 & The robot's verbal outputs were natural and contextually appropriate. & $\square$ & $\square$ & $\square$ & $\square$ & $\square$  \\
\hline
Q5 & The robot adapted appropriately to the patient’s affective state. & $\square$ & $\square$ & $\square$ & $\square$ & $\square$ \\
\hline
Q6 & The dialogue content reflected the patient’s background or memories. & $\square$ & $\square$ & $\square$ & $\square$ & $\square$  \\
\hline
Q7 & The system operated smoothly without noticeable technical issues. & $\square$ & $\square$ & $\square$ & $\square$ & $\square$ \\
\hline
Q8 & The patient showed signs of increased willingness to communicate. & $\square$ & $\square$ & $\square$ & $\square$ & $\square$  \\
\hline
Q9 & I would recommend this system for future reminiscence activity use. & $\square$ & $\square$ & $\square$ & $\square$ & $\square$  \\
\hline
\end{tabular}

\vspace{0.5cm}
Q1. Describe any notable patient reactions or emotional shifts you observed during the interaction.\vspace{0.5cm} \\
\vspace{0.5cm}\rule{\columnwidth}{0.3pt} \\ \rule{\columnwidth}{0.3pt}

Q2. Do you think the system had a positive impact on the patient's mood or social engagement? Please elaborate.\vspace{0.5cm} \\
\vspace{0.5cm}\rule{\columnwidth}{0.3pt} \\ \rule{\columnwidth}{0.3pt}

Q3. Do you have any suggestions for improving the system in future deployments?\vspace{0.5cm} \\
\vspace{0.5cm}\rule{\columnwidth}{0.3pt} \\ \rule{\columnwidth}{0.3pt}
}


\bibliography{IEEEabrv, bib/paper}
\bibliographystyle{IEEEtran} 
%

\newpage

 




\vfill

\end{document}